\documentclass[sigconf]{aamas} 

\usepackage{balance} 
\usepackage{amsmath,amsfonts}
\usepackage{bbm}
\usepackage{color}
\usepackage{tabto}

\usepackage{pdfpages}

\setcopyright{ifaamas}
\acmConference[AAMAS '22]{Proc.\@ of the 21st International Conference
on Autonomous Agents and Multiagent Systems (AAMAS 2022)}{May 9--13, 2022}
{Online}{P.~Faliszewski, V.~Mascardi, C.~Pelachaud,
M.E.~Taylor (eds.)}
\copyrightyear{2022}
\acmYear{2022}
\acmDOI{}
\acmPrice{}
\acmISBN{}

\acmSubmissionID{65}

\title[Interpretable PbRL]{Interpretable Preference-based Reinforcement Learning\\with Tree-Structured Reward Functions}

\author{Tom Bewley}
\affiliation{
  \institution{University of Bristol}
  \city{Bristol}
  \country{United Kingdom}
  }
\email{tom.bewley@bristol.ac.uk}

\author{Freddy Lecue}
\affiliation{
  \institution{CortAIx, Thales}
  \city{Montr\'eal}
  \country{Canada}}
\email{freddy.lecue@inria.fr}

\begin{abstract}
The potential of reinforcement learning (RL) to deliver aligned and performant agents is partially bottlenecked by the reward engineering problem. One alternative to heuristic trial-and-error is preference-based RL (PbRL), where a reward function is inferred from sparse human feedback. However, prior PbRL methods lack interpretability of the learned reward structure, which hampers the ability to assess robustness and alignment. We propose an online, active preference learning algorithm that constructs reward functions with the intrinsically interpretable, compositional structure of a tree. Using both synthetic and human-provided feedback, we demonstrate sample-efficient learning of tree-structured reward functions in several environments, then harness the enhanced interpretability to explore and debug for alignment.
\end{abstract}

\newcommand{\BibTeX}{\rm B\kern-.05em{\sc i\kern-.025em b}\kern-.08em\TeX}

\begin{document}


\pagestyle{fancy}
\fancyhead{}


\maketitle 

\section{Introduction}

It has recently been argued that the paradigm of reinforcement learning (RL) 
, in which agents learn action-selection policies by exploration to maximise future reward, is sufficiently general to give rise to most, if not all, aspects of natural and artificial intelligence \cite{silver2021reward}. However, the origin of the reward signal itself has received limited research attention compared with the algorithms used to optimise it. \citet{singh2009rewards} argue that the traditional presentation of reward, as a known and hard-wired property of an agent's environment, is misleading with respect to both biological plausibility and real-world technical applications. Motivated by evolutionary biology, they propose a differentiation between a latent fitness function $F$, which produces a scalar evaluation of the true quality of a behavioural trajectory, and the reward function $R$, which is a mere proxy to convert global pressures on fitness into local pressures on immediate decision-making. $R$ may have a nontrivial relationship to $F$ because it must be tailored to both the learning dynamics of the agent and the structure of the environment. During the learning process, two nested optimisation loops are at work: inner adaptation of the RL agent within a given reward structure $R$, and outer adaptation of $R$ to better align it with fitness.

As RL becomes more powerful, the engineering of aligned reward functions will become ``both more important and more difficult'' \cite{dewey2014reinforcement}. The prevailing reliance on heuristic approaches, which already presents challenges to real-world deployment \cite{leike2018scalable} and hampers the use of RL non-experts \cite{wirth2017survey}, is likely to prove inadequate in the face of vastly more capable agents which can exploit any specification errors. Several alternatives to heuristic reward engineering have been proposed, including inverse RL \cite{abbeel2004apprenticeship}, inverse reward design \cite{NIPS2017_32fdab65}, and advice-taking mechanisms \cite{maclin1996creating}, all of which involve interaction with a human with (possibly tacit) knowledge of $F$. In this paper, we focus on yet another human-in-the-loop approach, \textit{preference-based RL} (PbRL) \cite{wirth2017survey}, in which a reward function is inferred from preferences expressed by the human over sets of candidate behaviours, indicating which have higher fitness. Given a dataset of preference labels, $R$ is constructed to reward commonly-preferred states and actions, and this function is used to train the agent. While ostensibly restrictive, preference feedback places low cognitive demands on the human, requires little domain expertise or training, and tends to yield lower variance than direct fitness labelling since it does not require the human to fix an absolute scale \cite{guo2018experimental}.  

The outcome of PbRL is a policy whose true fitness depends on its expected reward under $R$, and the alignment of $R$ to $F$. Without ground-truth knowledge of $F$, it is hard to define quantitative metrics for the latter, which instead becomes a fuzzy, multi-faceted judgement, requiring the assessor to build an intuitive understanding of the structure of $R$ and its effect on agent learning. In large part, this is an \textit{interpretability} problem. The importance of interpretability for human-in-the-loop RL has been highlighted in surveys \cite{leike2018scalable,arzate2020survey}, and some post hoc analysis has been applied to learnt reward functions to gain some insight into feature influence \cite{russell2019explaining,michaud2020understanding}, but to our knowledge, there have been no efforts to make $R$ \textit{intrinsically} interpretable (loosely speaking, human-readable) by constraining its functional form. Indeed, prior works implement reward functions as deep neural networks, or ensembles thereof \cite{christiano2017deep}, which are notoriously opaque to scrutiny. 

We present a PbRL algorithm that learns intrinsically interpretable reward functions from human preferences. Specifically, it yields \textit{tree-structured} reward functions, formed of independent components associated with disjoint subsets of the state-action space, and defined hierarchically as a binary tree. The tree is incrementally refined as new preference labels arrive, and the traceability of these changes provides a powerful mechanism for monitoring and debugging. Trees afford both diagrammatic and geometric visualisation, textual description as a rule set in disjunctive normal form, and the efficient computation of feature importance metrics. Maintaining an explicit uncertainty estimate for each reward component also facilitates active preference learning based on upper confidence bounds. We evaluate our algorithm on four benchmark RL problems using both synthetic and human feedback, and in both offline and online learning settings. We observe aligned and sample-efficient learning of tree-structured reward functions in each of these contexts, alongside some informative failure cases. We then harness the enhanced interpretability to explore and debug for alignment.

This paper is structured as follows. Section \ref{sec:prob_def} reviews the general PbRL problem definition, section \ref{sec:constraint} introduces the additional tree-structuring constraint that we apply to improve interpretability, and section \ref{sec:method} presents our algorithm for approximately solving the constrained problem. Section \ref{sec:res_perf} gives experimental results with a focus on quantitative performance metrics, while section \ref{sec:res_interp} uses case studies to demonstrate the qualitative interpretability benefits of the tree structure for the purpose of alignment. Finally, section \ref{sec:conc} briefly concludes and discusses directions for future work.

\section{PbRL Problem Definition} \label{sec:prob_def}

The PbRL problem is formalised within a Markov Decision Process without reward (MDP\textbackslash R) \cite{abbeel2004apprenticeship}, in which at discrete time $t$, an agent's action $a_t\in\mathcal{A}$ influences the evolution of an environmental state $s_t\in\mathcal{S}$ according to a Markovian dynamics function $D(s_{t+1}|s_t,a_t)$. We specifically consider fixed-length episodic MDP\textbackslash Rs, in which $t$ is initially $0$, $s_0$ is sampled from an initial distribution $P_0$, and the process deterministically terminates at a fixed $t=T$. The events of an episode are described by a trajectory in state-action space, $\tau=((s_{0},a_{0}),...,(s_{{T-1}},a_{{T-1}}))\in(\mathcal{S}\times\mathcal{A})^{T}$. There also exists a human observer, who evaluates the quality of trajectories according to the latent fitness function $F:(\mathcal{S}\times\mathcal{A})^T\rightarrow\mathbb{R}$. The ultimate goal of the agent is to learn an action selection policy $\pi(a_t|s_t)$ that maximises the expected fitness over induced trajectories:
\begin{equation} \label{eq:obj_max_fitness}
    \underset{\pi}{\text{argmax}}\ \mathbb{E}_{\tau_i\sim \text{Pr}(\tau|P_0,D,\pi)}F(\tau_i).
\end{equation}

In order to learn about $F$, the agent must interact with the human. In PbRL, we assume the human cannot specify the analytical form of $F$, or even evaluate it absolutely for a given trajectory, but can only assess the \textit{relative} fitness of a trajectory pair $\tau_i,\tau_j$ and provide a label $y_{ij}\in [\varepsilon,1-\varepsilon]$ indicating their assessment of the probability that $\tau_i$ has higher fitness than $\tau_j$ (denoted by $\tau_i\succ \tau_j$). Here, $\varepsilon\in(0,0.5]$ is a noise parameter preventing extreme probabilities. Agent-human interaction therefore consists of sampling trajectory pairs from a distribution $\psi:((\mathcal{S}\times\mathcal{A})^T)^2\rightarrow[0,1]$ and obtaining preference labels in response. The inference of $F$ reduces to minimising some loss $\ell$ over labelled pairs:
\begin{equation} \label{eq:obj_min_loss}
\underset{F}{\text{argmin}}\ \mathbb{E}_{(\tau_i,\tau_j)\sim \psi}\ \ell(y_{ij},P(\tau_i\succ \tau_j|F)).
\end{equation}
Here, $P$ is a statistical model of human preference labels given estimated fitness values for a trajectory pair, whose definition is an assumption of the modelling process. Applying a basic rationality hypothesis, we can assume that the probability of the human preferring $\tau_i$ to $\tau_j$ is a monotonically non-decreasing function of the fitness difference $F(\tau_i)-F(\tau_j)$. Specifically, we adopt Thurstone's law of comparative judgement \cite{thurstone1927law}, which models the fitness of a set of $n$ trajectories $F(\tau_1),...,F(\tau_n)$ as a multivariate normal distribution with mean $\boldsymbol{\mu}\in\mathbb{R}^n$ and covariance $C\in\mathbb{R}^{n\times n}$.\footnote{Prior works \cite{christiano2017deep,xu2020preference} have adopted the Bradley-Terry model \cite{bradley1952rank}  which maintains no covariance estimate and uses a logistic function $P(\tau_i\succ \tau_j|\boldsymbol{\mu})=\frac{1}{1+\exp(\boldsymbol{\mu}_j-\boldsymbol{\mu}_i)}$ . Both models are well-established in the preference modelling literature and often behave similarly in practice, but Thurstone's better matches the statistical assumptions of our method. As we explore later, the estimation of $C$ also provides a natural mechanism for uncertainty-driven active learning.} This leads to the following preference model:
\begin{equation} \label{eq:obj_min_loss_decomp}
P(\tau_i\succ \tau_j|\boldsymbol{\mu},C)=\Phi\left(\frac{\boldsymbol{\mu}_i-\boldsymbol{\mu}_j}{\sqrt{C_{ii}+C_{jj}-2C_{ij}}}\right),
\end{equation}
where $\Phi$ is the standard normal cumulative distribution (CDF). 

Following the classic approach of \citet{mosteller1951remarks}, we note that under Thurstone's model, a label $y_{ij}$ implies that the variance-scaled difference in fitness between $\tau_i$ and $\tau_j$ is proportional to $\Phi^{-1}(y_{ij})$, where $\Phi^{-1}$ is the inverse normal CDF.\footnote{The importance of $\varepsilon$-bounding $y_{ij}$ is revealed here: it ensures that applying $\Phi^{-1}$ cannot yield infinite values.} Therefore, a suitable choice for the labelling loss $\ell$ is the squared error in this variance-scaled fitness difference. Equation \ref{eq:obj_min_loss} can be rewritten as 
\begin{equation} \label{eq:obj_min_loss_lstsq}
\underset{\boldsymbol{\mu},C}{\text{argmin}}\ \mathbb{E}_{(\tau_i,\tau_j)\sim \psi}
\left[\Phi^{-1}(y_{ij})-\frac{\boldsymbol{\mu}_i-\boldsymbol{\mu}_j}{\sqrt{C_{ii}+C_{jj}-2C_{ij}}}\right]^2.
\end{equation}

For the agent to perform the optimisation in equation \ref{eq:obj_max_fitness}, $\boldsymbol{\mu}$ and $C$ must be parameterised in a way that generalises to unlabelled trajectories. 
We adopt a linear model $\boldsymbol{\mu}_i=\textbf{r}^\top\textbf{n}_i,\ C_{ij}=\textbf{n}_i^\top\Sigma\textbf{n}_j,\ \forall i,j\in\{1..n\}$, where $\textbf{n}_i\in\mathbb{R}^m$ is a feature vector summarising the trajectory, $\textbf{r}\in\mathbb{R}^m$ is a vector of weights, and $\Sigma\in\mathbb{R}^{m\times m}$ is a covariance matrix associated with $\textbf{r}$. As in several prior works \cite{abbeel2004apprenticeship,akrour2012april,christiano2017deep}, we add a second level of decomposition by defining $\phi:\mathcal{S}\times\mathcal{A}\rightarrow\mathbb{R}^m$ as a function that constructs feature vectors from \textit{individual state-action pairs}, and $\textbf{n}_i=\sum_{t=0}^{T-1}\phi(s_{it},a_{it})$ as the unweighted feature expectation over $\tau_i$.\footnote{See Appendix A for a brief discussion of the psychological assumptions underlying the feature expectation decomposition.} By the linearity of the normal distribution, it follows that $\boldsymbol{\mu}_i-\boldsymbol{\mu}_j=\textbf{r}^\top(\textbf{n}_i-\textbf{n}_j)$ and $C_{ii}+C_{jj}-2C_{ij}=
(\textbf{n}_i-\textbf{n}_j)^\top \Sigma(\textbf{n}_i-\textbf{n}_j)$. The final form of equation \ref{eq:obj_min_loss} is thus
\begin{equation} \label{eq:obj_min_loss_lstsq_decomp}
\underset{\phi,\textbf{r},\Sigma}{\text{argmin}}\ \mathbb{E}_{(\tau_i,\tau_j)\sim \psi}
\left[\Phi^{-1}(y_{ij})-\frac{\textbf{r}^\top(\textbf{n}_i-\textbf{n}_j)}{\sqrt{(\textbf{n}_i-\textbf{n}_j)^\top\Sigma(\textbf{n}_i-\textbf{n}_j)}}\right]^2.
\end{equation}

The decomposition also allows us to rewrite equation \ref{eq:obj_max_fitness} as
\begin{equation} \label{eq:obj_max_fitness_decomp}
    \underset{\pi}{\text{argmax}}\ \mathbb{E}_{\tau_i\sim \text{Pr}(\tau|P_0,D,\pi)}\sum_{t=0}^{T-1}\textbf{r}^\top\phi(s_{it},a_{it}),
\end{equation}
which is structurally identical to the conventional RL objective of maximising (undiscounted) return in an MDP with reward. Thus, once the agent has inferred the function $\phi$ and vector $\textbf{r}$ it can define a reward function $R(s,a)=\textbf{r}^\top\phi(s,a)$, then employ any unmodified RL algorithm to learn a policy $\pi$. As alluded to in the Introduction, $R$ thereby serves as a proxy for $F$, with the true fitness of $\pi$ being a function of both its expected return under $R$ (equation \ref{eq:obj_max_fitness_decomp}) and the alignment of $R$ to $F$, which is approximated by the labelling loss (equation \ref{eq:obj_min_loss_lstsq_decomp}). Given that equation \ref{eq:obj_max_fitness_decomp} is the domain of standard RL, the contribution of this paper is an algorithm for approximately solving equation \ref{eq:obj_min_loss_lstsq_decomp}, subject to the particular interpretability constraint outlined in the following section.

\section{Interpretability Constraint} \label{sec:constraint}

\begin{figure}[H]
\includegraphics[width=\columnwidth ,interpolate=false]{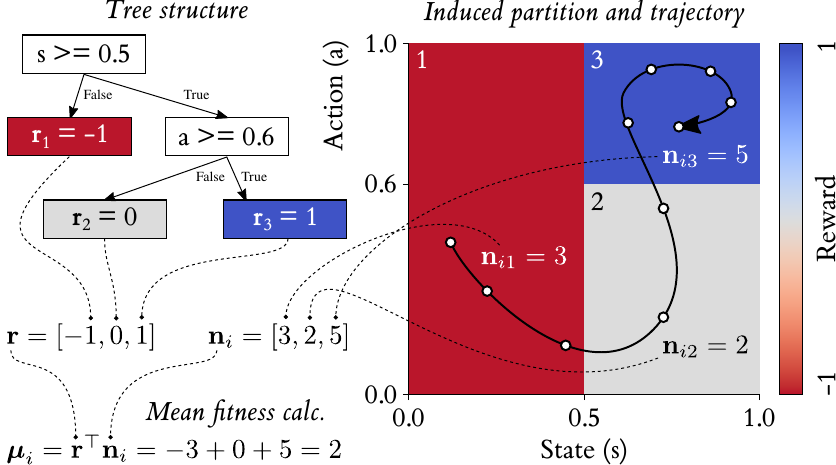}
\caption{Tree and induced partition for a simple case with $D_s=D_a=1$ and $m=3$. For a trajectory $\tau_i$, $\textbf{n}_i$ counts the timesteps (white circles) spent in each subset, and its dot product with $\textbf{r}$ gives the mean fitness estimate $\boldsymbol{\mu}_i$. Each component also has a variance $\Sigma_{xx}$, which can be used to compute the trajectory fitness variance $C_{ii}=n_i\Sigma n_i$ (not shown).
}
\label{fig:constraint_diagram}
\end{figure}

We now introduce the key assumption that differentiates our approach from prior work and enables interpretability. That is, we constrain the feature function $\phi$ so that for all $(s,a)\in\mathcal{S}\times\mathcal{A}$, $\phi(s,a)$ is a one-hot vector. This effectively induces a partition of the state-action space into $m$ disjoint subsets, which map to the $m$ possible one-hot vectors. For trajectory $\tau_i$, $\textbf{n}_i$ can be interpreted as the number of timesteps spent in each subset, and the reward vector $\textbf{r}$ as a set of \textit{components} that reward state-action pairs according to the subsets they fall within. We model reward components as independent, so that $\Sigma$ is a diagonal matrix. Furthermore, the partition induced by $\phi$ has a \textit{binary tree} structure, with the $m$ subsets as leaves, connected by a hierarchy of internal nodes emanating from a root. Each internal node applies a test to the state-action pair $(s,a)$. If the test is passed, the logical flow proceeds to the ``right'' child node. Otherwise, it proceeds to the ``left'' child. Testing continues until a leaf node $x\in\{1..m\}$ is reached, the state-action pair is mapped to the corresponding one-hot vector, and the reward is given as the corresponding component $\textbf{r}_x$ with variance $\Sigma_{xx}$. Although most of our algorithm does not rely on this assumption, we focus here on Euclidean state and action spaces $\mathcal{S}=\mathbb{R}^{D_s}$, $\mathcal{A}=\mathbb{R}^{D_a}$, in which a state-action pair is a vector $(s,a)\in\mathbb{R}^{D}$, where $D=D_s+D_a$. Internal node tests have the form $(s,a)_d\geq c:d\in \{1..D\}$, which evaluates whether the $d$th element of $(s,a)$ meets or exceeds a threshold $c$. Consequently, each state-action subset has the geometry of an axis-aligned hyperrectangle. Figure \ref{fig:constraint_diagram} provides an illustrative example.

\section{Proposed Algorithm} \label{sec:method}

We now present an algorithm for optimising equation \ref{eq:obj_min_loss_lstsq_decomp} subject to the tree-structuring constraint on $\phi$, alongside an active learning scheme that adapts the sampling distribution $\psi$ to focus on trajectories with high-variance fitness estimates and correct overestimation errors. As the algorithm iterates over stages of preference elicitation, reward component fitting, tree structure refinement and distribution updates, there is no single first step. We have tried to order the following subsections to maximise comprehensibility.

\paragraph{Remark} The indirect, multi-stage optimisation approach described here was converged on after extensive experimentation with alternatives, which we outline in Appendix B. The final approach is computationally efficient, easy to implement, and yields reward functions that are significantly more robust to small data changes than the alternatives.

\subsection{Preference Elicitation and Representation}

We first outline the process of sampling trajectory pairs and storing preference labels. We assume a finite data setting, in which the trajectory space $(\mathcal{S}\times\mathcal{A})^T$ is approximated by a sequence of $n$ trajectories $\mathcal{T}=(\tau_1,...,\tau_n)$, and the domain of $\psi$ is restricted to $\mathcal{T}^2$. This distribution can thus be written as a matrix $\Psi\in[0,1]^{n\times n}:\sum_{ij}\Psi_{ij}=1,\Psi_{ii}=0,\ \forall i\in\{1..n\}$. For the moment, let us take both $\mathcal{T}$ and $\Psi$ as given; we discuss their origins in section \ref{sec:sampling}. An instance of preference feedback is obtained by sampling a trajectory pair $\tau_i$, $\tau_j$ with probability $\Psi_{ij}$, presenting the pair (e.g. by visualisation) to a human, and recording the resultant preference label $y_{ij}\in[\varepsilon,1-\varepsilon]$. We represent a set of $k$ feedback instances by three data structures, a set $\mathcal{P}$, a matrix $A\in\{-1,0,1\}^{k\times n}$ and a vector $\textbf{y}\in\mathbb{R}^k$, which are incrementally assembled as preferences arrive. After sampling the $k$th pair $\tau_i,\tau_j$ and observing $y_{ij}$, we add $\{\tau_i,\tau_j\}$ to $\mathcal{P}$, append a row to $A$ in which the $i$th element is $1$, the $j$th element is $-1$ and all other elements are $0$, and append $\textbf{y}_k=y_{ij}$. $\mathcal{P}$ thus serves as a record of which pairs have been sampled. $A$ and $\textbf{y}$ allow us to express equation \ref{eq:obj_min_loss_lstsq_decomp} in matrix form as follows:
\begin{equation} \label{eq:obj_min_loss_lstsq_matrix}
\underset{\phi,\textbf{r},\Sigma}{\text{argmin}}\ \left[\Phi^{-1}(\textbf{y})-(\text{diag}(N A^\top\Sigma AN^\top)^{-\frac{1}{2}})^\top AN^\top\textbf{r}\right]^2.
\end{equation}
Here, $N$ is the $m\times n$ matrix of columnar feature vectors $\textbf{n}_i$. This notation is used in equations throughout this section.

\vspace{-0.25cm}
\subsection{Trajectory-Level Fitness Estimation}

Direct optimisation of equation \ref{eq:obj_min_loss_lstsq_matrix} subject to the tree-structuring constraint is computationally intractable, so we approximate the global problem by a sequence of local ones. First, we temporarily apply Thurstone's \textit{Case V} reduction \cite{thurstone1927law}, which assumes unit standard deviations for all fitness differences, i.e. $\text{diag}(N A^\top\Sigma AN^\top)^{\frac{1}{2}}=\mathbbm{1}_m$, so simplifies the bracketed expression in equation \ref{eq:obj_min_loss_lstsq_matrix} to $\Phi^{-1}(\textbf{y})-AN^\top\textbf{r}$. Following the method developed by Morrissey and \citet{gulliksen1956least}, we then use least squares to compute a somewhat biased (by the Case V assumption) estimate of $N^\top\textbf{r}=\boldsymbol{\mu}$, which is the $n\times 1$ vector of mean fitness estimates at the level of complete trajectories:
\begin{equation} \label{eq:morrissey_gulliksen}
N^\top\textbf{r}=\boldsymbol{\mu}\approx\tilde{\boldsymbol{\mu}}=(A^\top A)^{-1}A^\top\Phi^{-1}(\textbf{y}).
\end{equation}

\vspace{-0.4cm}
\subsection{Independent Reward Component Fitting} \label{sec:components}

We then use the trajectory-level estimates $\tilde{\boldsymbol{\mu}}$ to fit the means and variances of the reward components. This is a kind of temporal credit assignment problem: how does each component contribute to the overall fitness of a trajectory $\tau_i$? To solve it, we recall that under our tree-structuring constraint, $N_{xi}$ is the number of timesteps $\tau_i$ spends in the $x$th state-action subset. A priori, we cannot know which timesteps are responsible for the fitness of $\tau_i$, so the least strong assumption is a uniform temporal prior, that the reward component for this subset contributes a fraction $N_{xi}/T$ of the fitness. We can thereby compute an empirical estimate of the $x$th reward component mean by taking a weighted sum over trajectories:
\begin{equation} \label{eq:fit_comp_mean}
    \textbf{r}_x=\frac{\sum_{\tau_i\in\mathcal{P}}\frac{N_{xi}}{T}\tilde{\boldsymbol{\mu}}_i}{\sum_{\tau_i\in\mathcal{P}}N_{xi}}.
\end{equation}
Note that this summation is only over trajectories for which at least one preference label has been provided (denoted, in a slight abuse of notation, by $\tau_i\in\mathcal{P}$). We then use $\tilde{\boldsymbol{\mu}}$ and $\textbf{r}_x$ to empirically estimate the variance of the $x$th component, $\Sigma_{xx}$:
\begin{equation} \label{eq:fit_comp_var}
    \Sigma_{xx}=\frac{\text{RSS}(N_x)}{\sum_{\tau_i\in\mathcal{P}}N_{xi}}; \ \ \ \ \   \text{RSS}(N_x)=\sum_{\tau_i\in\mathcal{P}}N_{xi}\left[\frac{\tilde{\boldsymbol{\mu}}_i}{T}-\textbf{r}_x\right]^2.
\end{equation}
The residual sum of squares $\text{RSS}(\cdot)$ is a useful intermediate function that we use again below. By this method, reward components are fitted independently, using only the corresponding rows of $N$. The independent treatment of components leaves $\Sigma$ as a diagonal matrix.

\vspace{-0.25cm}
\subsection{Tree Structure Refinement}
So far, we have assumed a fixed $N$ matrix, and thus a fixed feature function $\phi$, but this function can be modified by adding or removing internal nodes in the tree structure described above. Consider splitting the $x$th leaf node into two by replacing it with an internal node which tests whether the $d$th element of the $(s,a)$ vector exceeds a threshold $c$. The effect of this split on the feature matrix $N$ is to replace the $x$th row of $N$ with a new $2\times n$ matrix, denoted by $N^{[xdc]}$, representing how the number of timesteps each trajectory spends in the $x$th state-action subset is distributed between two new child subsets contained within it. Using the two rows of $N^{[xdc]}$, reward components for the child subsets can be fitted using equations \ref{eq:fit_comp_mean} and \ref{eq:fit_comp_var}. All other reward components remain unchanged. Repeatedly making such splits grows the tree, each time increasing the number of leaves, subsets and associated reward components $m$ by $1$.

Whenever the trajectory-level fitness estimates $\tilde{\boldsymbol{\mu}}$ are updated, our algorithm grows the existing tree by recursive splitting up to a maximum $m=m_\text{max}$, at each step choosing a component to split $x$, splitting dimension $d$ and threshold $c$ to greedily minimise the total residual sum of squares $\sum_{x=1}^m\text{RSS}(N_x)$, and thus achieve a better fit to $\tilde{\boldsymbol{\mu}}$. Since reward components are fitted independently, a single step of this optimisation process can be expressed as follows:
\begin{equation} \label{eq:split_criterion}
    \underset{1\leq x\leq m}{\text{max}}\ \underset{1\leq d\leq D}{\text{max}}\ \underset{c\in\mathcal{C}_{d}}{\text{max}}\ \text{RSS}(N_x)-\text{RSS}(N^{[xdc]}_1)-\text{RSS}(N^{[xdc]}_2).
\end{equation}
$\mathcal{C}_{d}$ is a set of candidate split thresholds along dimension $d$. In our experiments, where dataset sizes are moderate, we exhaustively search over all values that occur in labelled trajectories: $\mathcal{C}_{d}=\{(s,a)_d,\ \forall (s,a)\in\tau,\ \forall\tau\in\mathcal{P}\}$. Crucially, the RSS-based splitting criterion in equation \ref{eq:split_criterion} is precisely the one used in classical regression tree learning \cite{breiman2017classification}. Our algorithm thus utilises a virtually-unmodified, highly-optimised, regression tree implementation.

Once $m_\text{max}$ is reached, we then iterate backwards through the growth process, pruning the tree back until $m=1$. At each step in this backward pass, we use the corresponding $N$, $\textbf{r}$ and $\Sigma$ to evaluate the labelling loss expression given in equation \ref{eq:obj_min_loss_lstsq_matrix} (i.e. the global objective that we aim to optimise). To this labelling loss, we add a \textit{complexity regularisation} term $\alpha m$ to modulate the tradeoff between predictive accuracy and interpretability (through compactness) and also to mitigate overfitting. We identify the tree size that minimises the regularised labelling loss, and use this tree as the updated $\phi$.

\vspace{-0.25cm}
\subsection{Trajectory Pair Sampling Distribution} \label{sec:sampling}

Our algorithm works in two data settings: offline and online. In the offline setting, the underlying trajectory dataset $\mathcal{T}$ remains fixed but the sampling matrix $\Psi$ is modified over time. In the online setting, both $\Psi$ and $\mathcal{T}$ change, with the latter being gradually augmented with new trajectories. This makes it possible to use trajectory data generated by the PbRL agent itself as it learns a policy in real-time.

\paragraph{Offline setting} 

A wide variety of active preference learning schemes have been proposed for PbRL \cite{talati2021aprel}. Partly inspired by existing work in the bandit literature \cite{carpentier2011upper,zoghi2014relative} we adopt an upper confidence bound (UCB) strategy, which weights trajectory pairs according to optimistic estimates of their summed fitness. This strategy prioritises trajectories with highly uncertain fitness under the current model, for which additional preference labels are likely to be most useful for reducing uncertainty (in this respect, it is similar to \cite{christiano2017deep}). Additionally, the optimism induces a bias towards identifying and correcting cases where trajectory fitness is overestimated, ultimately yielding a conservative reward function which counteracts the well-known overestimation bias in value-based RL \cite{fujimoto2018addressing}. Finally, biasing the preference dataset towards promising trajectories leads the reward function to prioritise distinguishing between high and very high fitness behaviour (rather than low vs very low), which reduces the risk of an agent stagnating at mediocre fitness with no incentive to improve. To implement the UCB strategy we use $N$, $\textbf{r}$ and $\Sigma$ to compute a vector of optimistic fitness values:
\begin{equation} \label{eq:ucb}
    \textbf{u}=\boldsymbol{\mu}+\lambda\ \text{diag}(C)^{\frac{1}{2}}=N^\top\textbf{r}+\lambda\ \text{diag}(N^\top\Sigma N)^{\frac{1}{2}},
\end{equation}
where $\lambda\geq0$ determines the number of standard deviations added to the mean. We then construct an $n\times n$ weighting matrix as follows:
\begin{equation}
    W^\text{off}_{ij}=\left\{\begin{array}{ll}
    0&\text{if }i=j \text{ or }\{\tau_i,\tau_j\}\in \mathcal{P},\\
    &\text{or }(\mathcal{P}\neq\emptyset\text{ and }\tau_i\not\in \mathcal{P}),\\
    \textbf{u}_i+\textbf{u}_j+\delta &\text{otherwise,}
    \end{array}\right.
    \ \ 
    \begin{array}{ll}
    \forall i\in\{1..n\},\\
    \forall j\in\{1..n\}.
    \end{array}
\end{equation}
Here, the three ``zeroing'' conditions respectively prevent comparing a trajectory to itself, prevent repeated pairs, and ensure that one of any sampled pair has already received feedback.\footnote{The last condition ensures that the graph representing the set of pairwise comparisons $\mathcal{P}$ is connected, meaning there is a path between any two $\tau_i\in\mathcal{P}$, $\tau_j\in\mathcal{P}$. As shown by \citet{csato2015graph}, this is necessary for the least squares solution in equation \ref{eq:morrissey_gulliksen} to be unique.}
The offset $\delta$ is calibrated so that the minimum element not matching a zeroing condition is set to $0$.\footnote{Unless this is also the maximum element, in which case it is offset to a positive value (nominally $1$) to ensure that $\sum_{ij}W^\text{off}_{ij}>0$ and prevent a divide-by-zero error.} If \textit{all} elements match a zeroing condition, then all possible trajectory pairs have been sampled and the process must be halted. Otherwise, we define $\Psi=W^\text{off}/\sum_{ij}W^\text{off}_{ij}$.

\paragraph{Online setting}

If $\mathcal{T}$ monotonically expands with new trajectories over time, but preference labels are obtained at a constant rate, it is possible to show that a higher density of labels is given to trajectory pairs that appear earlier, creating a strong earliness bias in the preference dataset. Assuming a total labelling budget $k_\text{max}$ and known final trajectory count $n_\text{max}$, we correct for this bias by collecting a batch of labels every time $f_l$ new trajectories are added, using monotonically-increasing batch sizes. On the $b$th batch, we define $W^\text{on}$ the same as $W^\text{off}$, with the extra condition that $W^\text{on}_{ij}=0$ if $i\leq f_l(b-1)$ and $j\leq f_l(b-1)$, which ensures that at least one of $i$ and $j$ are in the most recent $f_l$ trajectories. We compute $\Psi$ by normalising $W^\text{on}$ as above, then sample $k_b$ trajectory pairs, where
\begin{equation} \label{eq:scheduling}
k_b=\text{round}\left(k_\text{max}\frac{f_l^2(2b-1)-f_l}{n_\text{max}(n_\text{max}-1)}\right).
\end{equation}
We refer the reader to Appendix C for a more detailed justification of this approach and a derivation of equation \ref{eq:scheduling}. Note that the offline setting is recovered by setting $f_l=n_\text{max}=|\mathcal{T}|$. 

\subsection{Complete Algorithm}

Our algorithm initiates with one reward component ($m=1$) and $\textbf{r}=[0]$, $\Sigma=[0]$. During batch $b$, labelling is paused every $f_u$ samples for an iteration of trajectory-level fitness estimation, reward component fitting and tree refinement. This leads to a modified $\textbf{u}$ vector, which alters the sampling matrix for the rest of the batch. In the online setting, where an RL agent uses the learnt reward to train a policy in real-time, the latest $\phi$ and $\textbf{r}$ are used as soon as an update is made. Otherwise, the algorithm runs until the budget $k_\text{max}$ is expended, and the final fixed reward is used to train a policy at a later time. A pseudocode algorithm is given in Appendix D.

\vspace{-0.25cm}
\section{Performance Results} \label{sec:res_perf}

\begin{figure*}
\includegraphics[width=\textwidth,interpolate=false]{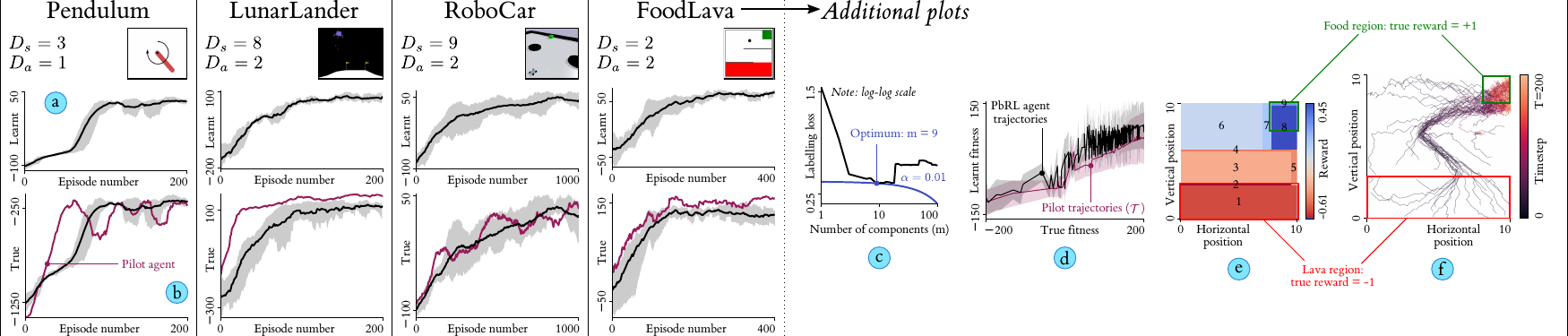}
\caption{Performance in offline setting using oracle feedback; additional plots shown for FoodLava.}
\vspace{-.05cm}
\label{fig:offline_oracle}
\end{figure*}
\begin{figure*}
\includegraphics[width=\textwidth,interpolate=false]{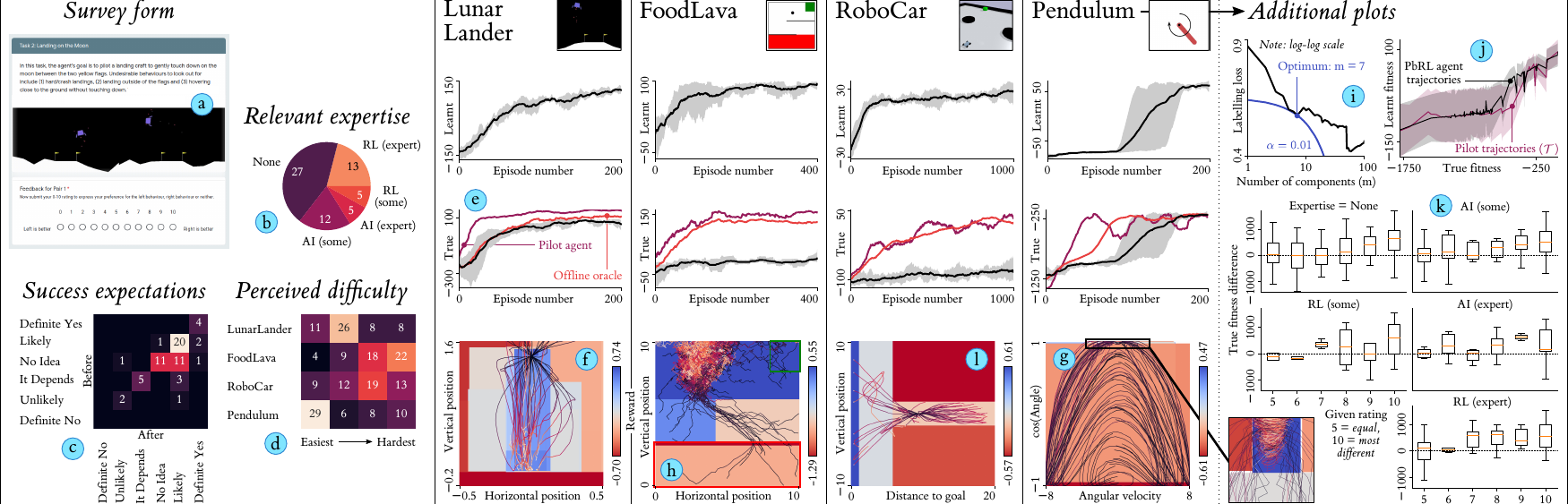}
\caption{Performance in offline setting using human feedback; additional plots shown for Pendulum.}
\vspace{-.05cm}
\label{fig:offline_human}
\end{figure*}
\begin{figure*}
\includegraphics[width=\textwidth,interpolate=false]{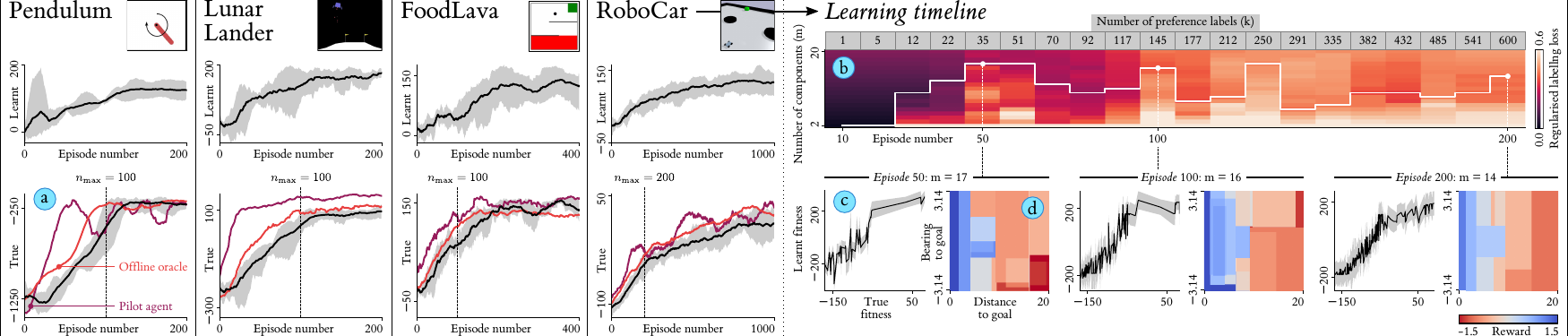}
\caption{Performance in online setting using oracle feedback; learning timeline shown for RoboCar.}
\vspace{-.05cm}
\label{fig:online_oracle}
\end{figure*}
\begin{figure*}
\includegraphics[width=\textwidth,interpolate=false]{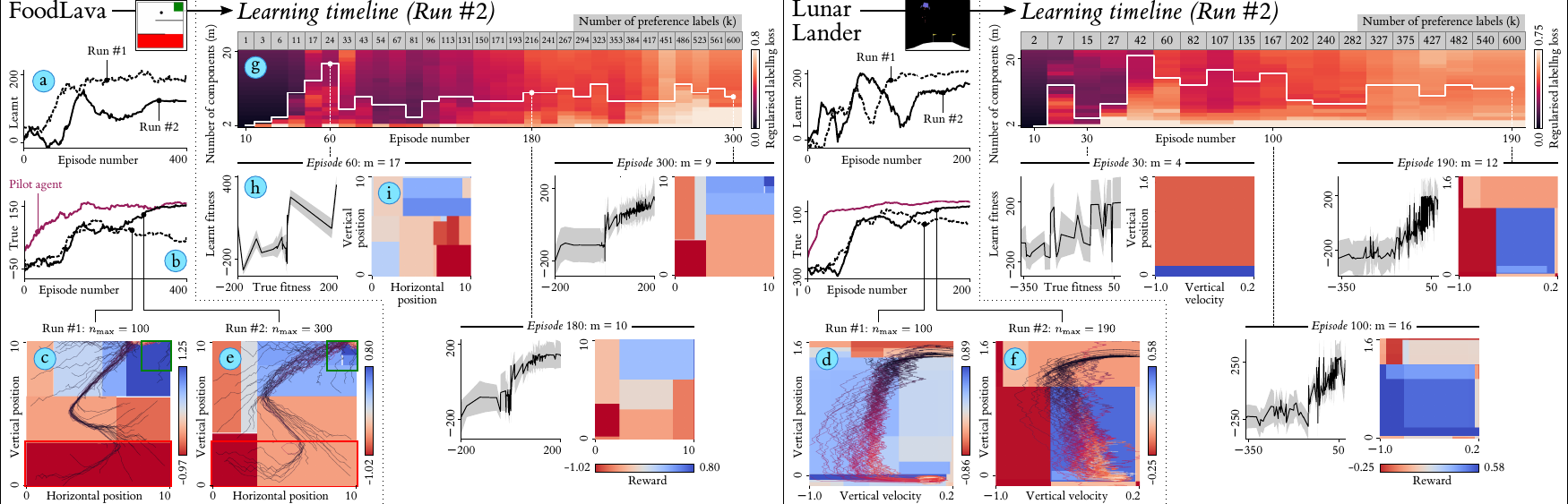}
\caption{Performance in online setting using human feedback; results with learning timelines for FoodLava and LunarLander.}
\label{fig:online_human}
\end{figure*}

We have evaluated our algorithm in four RL environments under various learning conditions. In all cases we used a feedback budget $k_\text{max}$ between $600$ and $620$, equating to $\approx 1$ hour of human time. Full experimental details are reported in Appendix E; below we discuss our findings by reference to the blue annotation letters in figures \ref{fig:offline_oracle}-\ref{fig:online_human}. Key findings are in \textbf{bold} and summarised at the end.

\vspace{-0.25cm}
\subsection{Offline with Oracle Feedback (Figure \ref{fig:offline_oracle})}

Here, $\mathcal{T}$ was fixed, and comprised of trajectories generated by an RL agent (the \textit{pilot agent}) as it trained on each environment's default, hand-engineered reward function. Preference labels came from a synthetic oracle with query access to this reward function. We then trained a second agent (the \textit{PbRL agent}) on the learnt reward, and finally measured the PbRL agent's alignment to the original reward function, which was taken to represent ground-truth fitness; see \cite{ibarz2018reward} for a similar method of quantitative evaluation. We show learning curves (time series of fitness per episode) for both learnt (a) and ground-truth (b) reward ($5$ repeats; mean and min-max range shown). The consistent monotonicity of the former indicates that tree-structured reward functions gave rise to stable agent learning in all cases, and the latter confirm that this learning was well-aligned with the ground-truth. \textbf{For Pendulum and RoboCar, asymptotic fitness was indistinguishable from the pilot, while for LunarLander and FoodLava, it was slightly below}. We show additional plots for FoodLava. (c) shows that the $\alpha m$-regularised labelling loss was minimised by a tree with $9$ leaves, hence $9$ reward components. (d) gives another measure of alignment by plotting true vs learnt fitness ($\pm 1$ std) for both the pilot trajectories $\mathcal{T}$ and those generated during PbRL agent training. In both cases there is a clear positive correlation, although the relationship for the latter is noisier, suggesting a degree of distributional shift. (e) visualises the $9$ reward components over the two state dimensions as coloured rectangles.\footnote{This and subsequent visualisations use a projection method introduced in \cite{bewley2021tripletree}, which represents the state-action subset for each component by its rectangular projection onto two dimensions, coloured by its mean reward. Where the projections of multiple subsets overlap, their colours are averaged, weighted by the number of samples falling within them in the trajectory set $\mathcal{T}$. In this first FoodLava visualisation, there are in fact no overlaps because the tree only contains splits along the two plotted dimensions.}
The components are arranged isomorphically to the maze layout, with negative reward in the red ``lava'' region and positive reward around the green ``food''. 
However, some misalignment is visible as high-reward component $8$ is too large in the vertical direction. Referring to (f), which plots the final $10$ trajectories of all $5$ PbRL repeats, we see that this misalignment led to policies that sometimes terminated just below the food. 

\vspace{-0.25cm}
\subsection{Offline with Human Feedback (Figure \ref{fig:offline_human})}

Using the same pilot trajectories as $\mathcal{T}$, we then gathered preference data from $62$ human participants via a survey (a) then trained PbRL agents using the resultant reward functions. We asked participants to indicate their level of relevant expertise (b) and expectations of the likelihood of our method succeeding, both before and after completing the survey (c), and to rank the tasks by perceived difficulty of giving feedback (d). Notably, this ranking turned out to predict the success of our method in this setting, since \textbf{for LunarLander and Pendulum we achieved asymptotic ground-truth fitness (e) only slightly below the oracle results}. Information about the ground-truth was not used anywhere in this experiment, which instead relied on participants' intuitive understanding of the tasks. The fact that it could nonetheless be well-optimised by the resultant PbRL agents indicates that for LunarLander and Pendulum, \textbf{human intuition was broadly aligned with the default reward functions}. Again, the coloured rectangle plots (f) and (g) provide insight into the learnt reward structure, with high reward in LunarLander given in a column above the landing zone, and in Pendulum given when the pole is upright with small angular velocity. The final $10$ trajectories from the PbRL runs are overlaid, showing that the agents sought out high-reward regions and consequently solved the respective landing and pole-balancing tasks. \textbf{We were unable to achieve aligned learning in FoodLava and RoboCar}, although in the former the outcome was not as catastrophic as the learning curve suggests. As can be seen in the rectangle/trajectory plot (h), the PbRL agent learnt to solve most of the maze but was not incentivised to proceed to the food, since a large positive reward component covered the entire upper third. This, we hypothesise, is evidence of a \textbf{causal confusion problem}: within the pilot run dataset, most trajectories that reached the upper third then went on to the food, so our uniform temporal credit assignment was unable to determine that the latter step was necessary for a favourable preference label. A similar issue arose in RoboCar, which we focus on in our interpretability analysis (section \ref{sec:res_interp}). For Pendulum, we include the labelling loss curve (i) and alignment plot (j), alongside box plots showing the agreement between provided preference labels and ground-truth fitness differences (k). Ratings from participants of all expertise levels generally aligned with fitness (above dotted line), with this trend becoming slightly more pronounced for more certain ratings (closer to $10$), and an indication that those with RL expertise exhibited somewhat lower variance than those without. 

\vspace{-0.25cm}
\subsection{Online with Oracle Feedback (Figure \ref{fig:online_oracle})}

Next, we deployed our algorithm in an online setting, using trajectories generated by the PbRL agent as it trained on the learnt reward in real-time. After $n_\text{max}$ trajectories were gathered, the reward function was fixed and the agent continued to train until convergence. Initially, we used synthetic oracle feedback. The ground-truth learning curves (a) indicate that overall performance was similar to the oracle-based offline setting, \textbf{with mean final fitness being slightly higher for Pendulum and FoodLava and slightly lower for LunarLander and RoboCar}. We further examine one of the five RoboCar runs via a hybrid visualisation that we call a \textit{learning timeline} (b). With $n_\text{max}=200$ and $f_l=10$ we had a total of $20$ labelling batches, over which the number of labels $k$ increased according to the scheduling equation \ref{eq:scheduling}. The heatmap shows how the regularised labelling loss varied as a function $m$ during the tree pruning sweep performed after each batch. The overlaid white curve shows how the tree size was modified accordingly to track the optimal $m$. As batches accumulated, the global pattern was that \textbf{$m$ first increased to a maximum, then remained somewhat below that maximum thereafter}, with large changes becoming less frequent. Inspecting the model at three checkpoints during training, we find that (c) \textbf{the positive correlation between true and learnt fitness became less noisy over time}, and (d) the reward components converged to an arrangement that positively rewarded both proximity to the goal and facing towards it (bearing $\approx 0$), doing so in an almost-symmetric manner.

\vspace{-0.25cm}
\subsection{Online with Human Feedback (Figure \ref{fig:online_human})}
Finally, we ran the algorithm online using feedback from a single human participant. Due to the labour-intensiveness of this experiment, we focused on two environments -- FoodLava and LunarLander -- both of which revealed the risk of prematurely fixing the reward structure. For both, we initially used $n_\text{max}=100$ as in the oracle-based experiments, and found the agent quickly converged to high fitness according to the learnt reward function (a; dotted lines) but after episode $n_\text{max}$, gradually lost fitness according to the ground-truth reward (b). This indicated that \textbf{the reward structure had been fixed in a state that was only partly aligned}, so that further optimisation hindered true performance. Using rectangle plots of the final reward components, we can see that for FoodLava (c), the maximum positive reward component was placed around the food region, but was ``loosely'' targeted as it exceeded the bounds of the food. The overlaid final trajectories show that the agent learnt to seek out this high reward, but sometimes stop short of the food itself. For LunarLander (d), a similarly ``loose'' reward function was learnt that gave high reward for a vertical position close to zero, regardless of the vertical velocity. The trajectory overlay indicates this reward function led the agent to maintain high negative velocity as it approached the ground, which was recognised as a crash landing by the ground-truth reward function. For both environments, we completed a second run using a higher $n_\text{max}$, thereby \textbf{distributing the same feedback budget over a larger fraction of the training process}, providing more time for the human to reactively fine-tune alignment. For these runs, we found that ground-truth fitness (b) \textbf{almost matched the pilot agent} from the offline oracle experiment, with \textbf{no sign of a performance dropoff}. For FoodLava, the rectangle plot of reward components (e) reveals a much smaller area of maximum reward that did not exceed the bounds of the food region and thus incentivised the agent to reliably enter it. For LunarLander (f), we had a very different reward structure to the first run, which gave positive reward for maintaining slow vertical velocity, and negative reward for exceeding a velocity threshold of $-0.55$, rather than merely rewarding reaching the ground. This incentivised the agent to gradually decelerate as its height decreased, resulting in softer landings that were not registered as crashes by the ground-truth reward function. For both environments, we show a learning timeline for the second run (g). We again see the trend of $m$ increasing to a maximum before stabilising at a lower value. The checkpoints also show the relationship between true and learnt fitness becoming less noisy over time (h) and several intermediate reward component layouts that emerged during training (i).

\vspace{-0.25cm}
\paragraph{Summary of Key Findings}

$\bullet\ $ With no exceptions, the tree-\hspace{-0.08cm} structured reward functions gave rise to stable, convergent reinforcement learning.
$\bullet\ $ Using several hundred instances of oracle feedback based on ground-truth reward functions, our algorithm could reconstruct those reward functions sufficiently well to train PbRL agents whose performance nearly matched that of conventional RL agents. $\bullet\ $ The aggregated preferences of $62$ human participants (offline), as well those of a single participant (online), yielded learnt reward functions that were similarly well-aligned to the ground-truth. This is despite participants having no direct knowledge of the ground-truth, instead relying on intuitive task understanding alone. $\bullet\ $ In the offline setting, dataset biases led to causal confusion, where the learnt reward incentivised state-action pairs that commonly appeared \textit{alongside} high-fitness behaviours, as well as the behaviours themselves. Careful rebalancing of training data, or moving to an online learning setup, would both help to mitigate this problem. $\bullet\ $ In the online setting, the main failure mode was ``loose'' alignment due to fixing the reward prematurely. Increasing $n_\text{max}$ gave more time to reactively correct for misaligned changes in behaviour. $\bullet\ $ In the online setting, the number of components $m$ tended to initially increase rapidly, then stabilise somewhat below the maximum later on, with large changes becoming less frequent. $\bullet\ $ In all human experiments, we did not encounter a single tree with a split along an action dimension, so that all rewards were a function of state only. We are wary to make a general claim about this result, but it is consistent with a recent suggestion that human teaching focuses on states over actions \cite{jauhri2020interactive}, and indicates future PbRL work may succeed by learning state-dependent rewards only.

\vspace{-0.15cm}
\section{Interpretability Demonstration} \label{sec:res_interp}

\begin{figure}[H]
\includegraphics[width=\columnwidth ,interpolate=false]{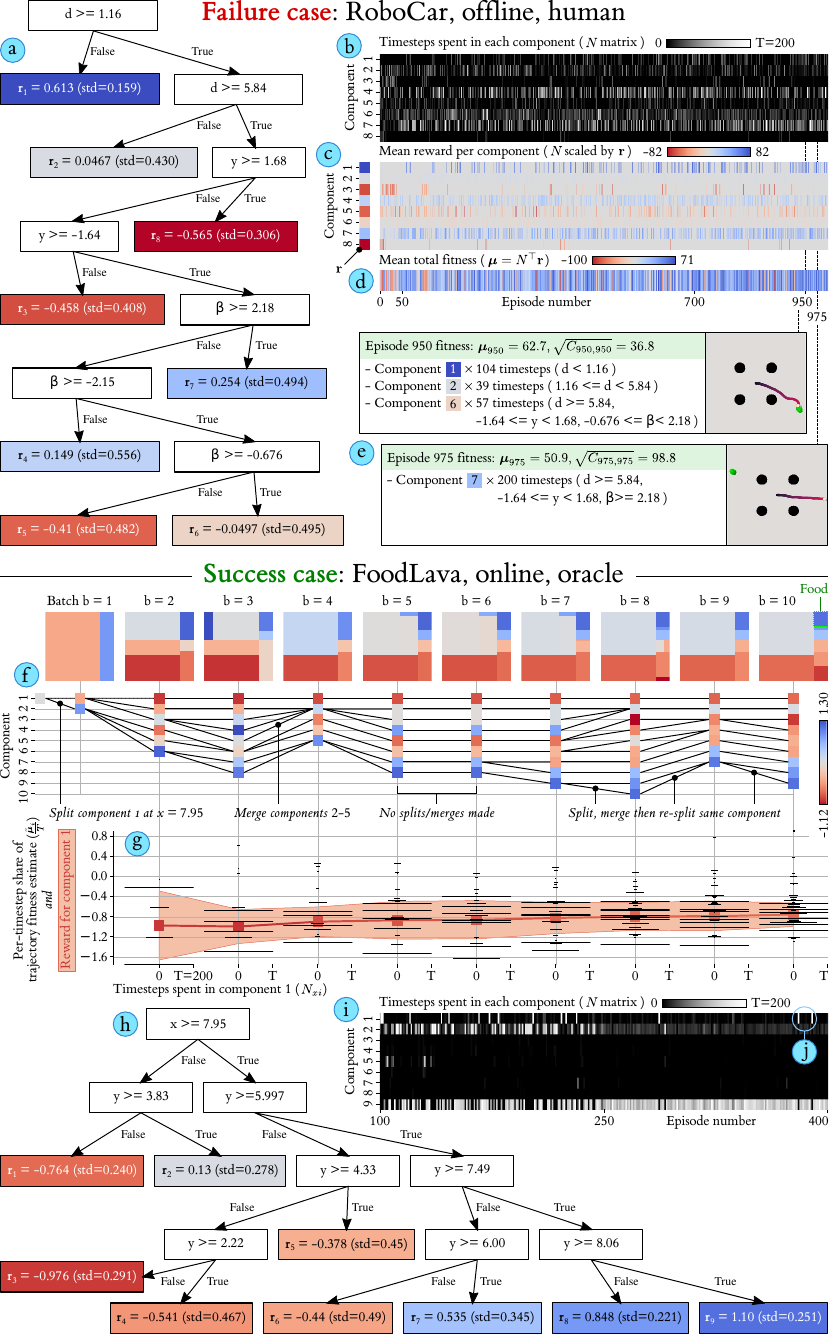}
\caption{Interpretability demo for two exemplar cases.}
\label{fig:interpretability}
\end{figure}

The analysis of rectangle plots above, which provides insight into the learnt reward functions and their effect on behaviour, exemplifies the interpretability benefits of the tree-structuring constraint. Figure \ref{fig:interpretability} demonstrates these benefits further by focusing on two specific PbRL runs: one failure case and one success case.

\vspace{-0.1cm}
\paragraph{Failure case: RoboCar using offline human feedback} As is visible in figure \ref{fig:offline_human} (l), this reward function led to policies that sometimes reached the goal as desired, but other times made no move towards the goal and appeared to seek only to maintain a vertical position close to $0$. We can diagnose this misalignment by examining the full reward function tree (figure \ref{fig:interpretability}; a). Here, the splitting dimensions are $y=$ vertical position (initialised to $0$), $d=$ distance to goal, and $\beta=$ bearing to goal in radians. The first two splits appear well-aligned, creating a component with maximum reward for achieving $d<1.16$ and a smaller positive reward for $d<5.84$. The remaining splits are problematic, creating components that penalise moving out of the region $y\in[-1.64,1.68]$ and, otherwise, reward a bearing outside of $\beta\in[-2.15,2.18]$ (i.e. facing \textit{away} from the goal). 
To understand this, consider the design of the environment. In each episode, the goal position is randomised but the car is initialised facing to the right, making it easier to reach the goal when it is also to the right. Hence, a majority of goal-reaching trajectories in the pilot run dataset showed the car driving directly forward, rarely exiting a narrow corridor around $y=0$. The splits to penalise large absolute $y$ are thus an example of causal confusion, in which behaviour correlating with a high fitness outcome is mistaken for being high fitness in itself, and would not appear if the environment were differently initialised or the dataset better balanced. We give a similar, if subtler, causal confusion justification of the $\beta$-based splits in Appendix F. The heatmaps (b), (c) and (d) provide fine-grained insight into the effect of the misaligned reward function on the learning dynamics of one of the $5$ PbRL repeats. (b) represents the timesteps spent in each component -- the $N$ matrix -- over the $1000$-episode training history. Multiplying $N$ row-wise with the mean reward vector $\textbf{r}$ we obtain (c), which gives per-episode reward from each component, and can be understood as a \textit{decomposed learning curve}. Summing (c) column-wise gives (d), the total fitness for each episode, which is a conventional learning curve. From these, we find that the agent quickly (by episode $50$) learned to avoid negative-reward components $3$ and $8$, inducing an early bias towards exiting the $y\in[-1.64,1.68]$ corridor. With this bias in place, exploration was curtailed and the agent largely settled into the moderate positive rewards of components $4$ and $7$. Although there was a gradual increase in visitation to component $1$ (the one corresponding to reaching the goal) in the first half of training, the agent never completely prioritised this component, with visitation peaking around episode $700$ before dropping off again. In (e), we harness the component structure to construct textual \textit{report cards} for two episodes near the end of training ($950$ and $975$) which describe the state-action subsets that were visited. While both are in the top $10\%$ of episodes by performance on the learnt reward, the former is aligned (obtaining positive reward from components $1$ and $2$) while the latter is not (staying entirely in component $6$, thereby driving straight ahead despite the goal being behind it). 

\vspace{-0.19cm}
\paragraph{Success case: FoodLava using online oracle feedback} In this run (chosen randomly from the $5$ repeats), we achieved aligned learning. For the first $n_\text{max}=100$ episodes, a label batch was obtained at intervals of $f_l=10$ and the tree structure incrementally updated by growth and pruning. (f) depicts the net changes resulting from each update using both rectangle plots and a graph of the split/merge dependencies between components from batch-to-batch. Key events in the construction of the reward function included the merging of four components into one at $b=4$, and the corrective splitting, merging, and re-splitting (at a different threshold) of a component between $b=8$ and $b=10$, yielding the final maximum-reward component $9$ whose subset boundaries (horizontal position $x\geq 7.95$, vertical position $y\geq 8.06$) lined up almost exactly with the food region ($x\geq 8$, $y\geq 8$). From $b=2$ onwards, component $1$ was persistent, being neither split nor merged. However, its mean and variance were continually refined as preference labels arrived, a process that we visualise in (g). For each batch, every trajectory $\tau_i$ that had been labelled so far (of which there are more for later batches) is shown as a black horizontal line, whose vertical position corresponds to its trajectory-level fitness estimate $\tilde{\boldsymbol{\mu}}_i$ (divided by the episode length $T=200$) and whose width is proportional to the time spent in component $1$, $N_{1i}$. The method described in section \ref{sec:components} effectively fits a normal distribution to these lines, and the results for all batches are overlaid (mean as squares, $\pm1$ std as shading). Between batches $2$ and $10$, component $1$ saw a slight increase in its mean, and a gradual narrowing of its variance, as more trajectories were labelled. (h) depicts the fixed $9$-component tree used for the final $n_\text{post fix}=300$ training episodes, and (i) shows the timesteps spent in each component throughout this period. As of episode $100$, roughly equal time was being spent in components $2$ and $9$, but the latter came to dominate around episode $250$. Since component $9$ corresponded to the food region, this indicated consistent, rapid solution of the navigation problem, with occasional failures (j) attributable to the agent becoming stuck in component 1. 

\vspace{-0.15cm}

\vspace{-0.63cm}
\section{Conclusion} \label{sec:conc}
\vspace{-0.05cm}

We have presented an algorithm for interpretable PbRL using tree-structured reward functions, demonstrating successful learning of compact and aligned reward functions across four environments, alongside informative and actionable failure cases due to causal confusion in the offline setting and premature reward fixing in the online setting. We have also shown the value of interpretability for exploring and debugging the learnt reward structure. In the offline setting, future work could construct trajectory datasets using unsupervised agents that optimise for behavioural diversity (e.g. \cite{eysenbach2018diversity}) instead of our somewhat contrived pilot agents. In the online setting, there is scope for larger human experiments, with a focus on ablation and hyperparameter tuning. An additional layer of interpretability could be realised by integrating our method with agent architectures that learn decomposed value functions \cite{juozapaitis2019explainable}.

\clearpage

\bibliographystyle{ACM-Reference-Format} 
\bibliography{bibliography}

\clearpage


\section*{Supplementary material}

# APPENDIX A: DISCUSSION OF FEATURE EXPECTATION DECOMPOSITION

The feature expectation decomposition, in which a trajectory-level feature vector is constructed as a sum over constituent state-action pairs $\sum_{t=0}^{T-1} \phi(s_{it}, a_{it})$, was popularised by Abbeel and Ng [1] and has become a standard component of work on inverse RL, PbRL and imitation learning in the years since. In the PbRL context, it encodes the assumption that a human evaluates the fitness of a behavioural trajectory by independently assessing each state-action pair, then performing an accurate unweighted sum over the trajectory length $T$. The validity of this assumption depends on many contextual factors, including whether a pair of trajectories is presented to the human concurrently or consecutively. In the latter case, there is a possiblity for ordering to affect a human's judgement, including via the *anchoring* effect [13]. Assuming a visual mode of presentation, another important distinction is whether visualisations are static (e.g. line plots) or dynamic (e.g. videos). If trajectory presentation is dynamic, evaluation is a temporally-extended experience, and the *peak-end rule* indicates that the human may depart from a uniform treatment of timesteps and exhibit a bias towards later (thus better-remembered) parts of the trajectory, as well as those with the highest intensity of positive or negative valence [10]. In human experiments for this paper, we present trajectories as short videos, arranged side-by-side in an on-screen window, so biases from temporal ordering should not be in effect. We also have the videos loop infinitely and enforce no time limit on the feedback process, so participants are able to review all parts of each trajectory many times. We aim for this to partially mitigate the peak-end effect.

A more fundamental issue is that humans commonly reason in terms of counterfactuals and hypotheticals. There is thus a risk of preference judgements being made on the basis of ancitipated future actions, or outcomes that an agent *nearly* realised but didn't, rather than the actual contents of a given trajectory as assumed in the defining equations of PbRL. In a review of best practice for human-agent teaching, Knox et al. [11] hypothesise that overall judgements draw on a mixture of observed past behaviour (about which the feature expectation vector is informative) and anticipated future behaviour (about which the vector contains no information), but also note that "positive results" have been attained by methods which ignore the latter.

The proceeding discussion serves to highlight that rather than any well-evidenced psychological validity, the advantage of the feature expectation decomposition lies in its simplicity and mathematical convenience. It is required to derive many of the equations in the main paper, not least equation 6, which expresses fitness in terms of a reward function and enables the use of a conventional RL algorithm for policy learning. Given that these conveniences come at the cost of a potentially significant mis-modelling of the human judgement process, we see the construction of a more psychologically plausible fitness decomposition for PbRL as an important direction for future work.

# APPENDIX B: OTHER ATTEMPTED OPTIMISATION METHODS

Before converging on the optimisation approach given in the main paper (i.e. trajectory-level fitness estimation using Thurstone's Case V, uniform temporal credit assignment to reward components, RSS-based splitting criterion) we explored several less robust or efficient alternatives. We briefly describe these below.

*Bradley-Terry model for trajectory-level fitness estimation.* Prior to adopting Thurstone's Case V model for the calculation of $\tilde{\boldsymbol{\mu}}$, we tried using the somewhat more popular Bradley-Terry model [2], computing maximum likelihood fitness estimates using both the iterative minorisation-maximisation (MM) algorithm [9] and the newer random walk algorithm proposed by Negahban et al. [12]. While we found the latter to be significantly faster and thus better suited to an online learning setting, both tended to yield highly skewed fitness distributions, with most trajectories assigned a fitness very close to 0, while a small number of outliers had fitness estimates several orders of magnitude larger. We found that this was problematic when it came to computing the means and variances of reward components, as the outliers dominated any smaller distinctions between the rest of the trajectory set. In contrast, the least squares method using Thurstone's Case V gave a far more even and unskewed spread of fitness estimates (following a roughly normal distribution), so that no single trajectory had an outsized impact on the reward components. We also found that Bradley-Terry fitness estimates were less robust to individual labelling "mistakes" than the least squares Case V solutions, with a single incorrect preference label often leading to a dramatic reordering of the fitness ranking over the set of trajectories. We concluded that such high sensitivity would be undesirable, especially in human experiments where the chance of a mistake was high.

*Least squares regression for component fitting.* Separate from the method used to estimate $\tilde{\boldsymbol{\mu}}$ is the subsequent step of computing mean and variance estimates for individual reward components. Initially, we used the expression $N^\top \mathbf{r} \approx \tilde{\boldsymbol{\mu}}$ to frame the estimation of $\mathbf{r}$ as a second least squares regression problem: $\mathbf{r} \approx (NN^\top)^{-1}N\tilde{\boldsymbol{\mu}}$. This approach suffered from two major problems. Firstly, it tended to give brittle and poorly regularised results, with extremely high or low reward magnitudes assigned to state-action subsets that were only visited by especially high- or low-fitness trajectories (even if this was just for a single timestep) and near-zero reward for many others. Secondly, it fitted the reward components in a way that was not independent. This prevented our use of the highly-optimised classical regression tree algorithm for tree growth, and instead required a far more expensive algorithm that repeatedly computed a least squares solution for each candidate split threshold. Our final independent fitting method, based on the uniform temporal prior, solves both of these problems.

*TrueSkill for end-to-end component fitting.* We also tried bypassing the intermediate estimation of trajectory-level fitness $\tilde{\boldsymbol{\mu}}$ completely, instead translating directly from a set of preference labels $\{y_{ij}\}$ to reward component means and variances. For this, we used the TrueSkill algorithm [8], developed by Microsoft to assign skill scores to individual video game players on the basis of their history of team-level match results. We refer readers to the cited paper for algorithmic details. In our appropriation of TrueSkill, the "players" were the reward components, "skill scores" were reward values, "teams" were trajectories, "match results" were pairwise preference labels, and the algorithm's *partial play* feature was harnessed to

weight each component $x$'s contribution to a trajectory $\tau_i$ by the fraction of time spent in the $x$th state-action subset, $N_{xi}/T$ (mirroring the uniform temporal credit assignment used in our final algorithm). TrueSkill is also grounded in the equations of Thurstone's model, and estimates a mean and variance for each player's score. The mapping of features of the algorithm into our problem context was thus remarkably well-motivated, and we found that it produced robust and highly plausible reward estimates for a fixed set of reward components (i.e. fixed tree structure $\phi$). Crucially, however, the algorithm could not readily be adapted to handle the splitting or merging of reward components, as in our tree growth process, which effectively changes the number of players in a team. The only way of implementing this functionality would be to run the full algorithm from scratch whenever a change is made, which would induce a prohibitive runtime on the order of seconds for each split threshold considered during tree growth. This unfortunately meant that TrueSkill could not be used to solve the complete PbRL problem. We are nonetheless excited by the prospect of future work that uses TrueSkill as part of a preference learning framework, and are confident that the algorithm has productive applications far outside the domain of video gaming.

## APPENDIX C: JUSTIFICATION AND DERIVATION OF FEEDBACK SCHEDULING

Let $k_{\max}$ be the labelling budget and $n_{\max}$ be the final trajectory count. Trajectories are added to $\mathcal{T}$ one at a time, with $\tau_i : i \in \{1..n_{\max}\}$ denoting the $i$th trajectory added. In constant-rate labelling, we obtain $k_{\max}/n_{\max}$ new preference labels each time a trajectory is added. Ignoring the effect of UCB weighting, and the conditions on $\Psi$ that prevent duplicates and ensure connectivity,[1] the expected number of labels for a trajectory pair $\tau_i, \tau_j$ is

$$\rho(\tau_i, \tau_j) = \sum_{n=\max(i,j)}^{n_{\max}} \frac{k_{\max}/n_{\max}}{n(n-1)} = \frac{k_{\max}}{n_{\max}} \sum_{n=\max(i,j)}^{n_{\max}} \frac{1}{n^2 - n}$$
$$= \frac{k_{\max}}{n_{\max}} \frac{n_{\max} - (\max(i,j) - 1)}{n_{\max}(\max(i,j) - 1)} \propto \frac{n_{\max}}{\max(i,j) - 1} - 1, \quad (1)$$

which decreases as $\max(i, j)$ increases. Hence, earlier trajectory pairs obtain a higher density of labels. Our feedback scheduling method is designed to correct for this bias and produce a uniform label density prior to introducing UCB weighting and the conditions on $\Psi$. Concretely, we wish for the following to hold:

$$\rho(\tau_i, \tau_j) = \frac{k_{\max}}{n_{\max}(n_{\max} - 1)}, \quad \forall i, j \in \{1..n_{\max}\} : i \neq j. \quad (2)$$

To achieve this, we organise the labelling process into batches, with a batch collected every time $f_l$ new trajectories are added, where $n_{\max} \% f_l = 0$. Within the $b$th batch ($b \in \{1..n_{\max}/f_l\}$), we only permit the sampling of trajectory pairs for which at least one of $\tau_i$ and $\tau_j$ is in the most recent set of $f_l$ trajectories. This is practically achieved by adding the condition $W_{ij}^{\text{on}} = 0$ if $i \leq f_l(b-1)$ and $j \leq f_l(b-1)$. Crucially, each pair $\tau_i, \tau_j$ only meets this condition for exactly one batch, namely the $b$th, where $f_l(b-1) < \max(i,j) \leq f_l b$.

---

[1]We ignore the UCB weighting in this analysis because our aim is to achieve uniform label density *before* adding this effect. In theory, the duplication and connectivity conditions should be included, but this would greatly complicate the analysis and would not yield a closed-form expression for $k_b$.

The number of trajectory pairs that meet the condition for the $b$th batch, $z_b$, is the difference between the current total number of possible pairs, $f_l b(f_l b-1)$, and the number of pairs as of the previous batch, $f_l(b-1)(f_l(b-1) - 1)$:

$$z_b = f_l b(f_l b - 1) - f_l(b-1)(f_l(b-1) - 1) = f_l^2(2b-1) - f_l. \quad (3)$$

Let $k_b$ denote the size of the $b$th batch. Again ignoring the UCB weighting and constraints, we assume uniform sampling from the set of condition-matching trajectory pairs. The expected label density for $\tau_i, \tau_j$ can thus be written as

$$\rho(\tau_i, \tau_j) = \frac{k_b}{z_b} = \frac{k_b}{f_l^2(2b-1) - f_l}. \quad (4)$$

We now wish to find an expression for $k_b$ such that equation 2 holds. This can be done by equating equations 2 and 4

$$\rho(\tau_i, \tau_j) = \frac{k_{\max}}{n_{\max}(n_{\max} - 1)} = \frac{k_b}{f_l^2(2b-1) - f_l}, \quad (5)$$

and rearranging for $k_b$:

$$k_b = k_{\max} \frac{f_l^2(2b-1) - f_l}{n_{\max}(n_{\max} - 1)}. \quad (6)$$

This equation holds for all $b \in \{1..n_{\max}/f_l\}$, and thus can be used to *schedule* the rate of feedback so as to achieve uniform label density. Since in practice $k_b$ must be an integer, the equation given in the main paper includes a final round($\cdot$) operation.

## APPENDIX D: COMPLETE ALGORITHM

Algorithm 1 provides a complete pseudocode outline of our PbRL algorithm. It includes a subfunction, updateRewardFunction, which performs an iteration of trajectory-level fitness estimation, reward component fitting and tree structure refinement given the latest preference dataset, and is written separately in algorithm 2.

*Inputs.* Algorithm 1 takes the following as input:

- $\mathcal{T}_{\text{off}}$: Offline trajectory dataset. In the online setting $\mathcal{T}_{\text{off}} = []$.
- $n_{\max}$: Final size of trajectory dataset after collection of online data. In the offline setting, $n_{\max} = |\mathcal{T}_{\text{off}}|$.
- $f_l$: Batch frequency. In the offline setting, $f_l = n_{\max} = |\mathcal{T}_{\text{off}}|$.
- $\lambda$: Number of standard deviations added to $\mu$ to create optimistic fitness estimates $\mathbf{u}$ for UCB sampling.
- $k_{\max}$: Total labelling budget.
- $f_u$: Within-batch tree/reward function update frequency.
- $m_{\max}$: Maximum tree size allowed during growth stage.
- $D_s, D_a$: Dimensionality of state and action spaces.
- $\alpha$: Tree complexity regularisation parameter.
- $n_{\text{post fix}}$: Number of RL episodes after fixing reward function.

*Subfunctions.* In addition to updateRewardFunction, numerous other subfunctions are used. For brevity, we do not define these in detail, but describe them informally below, in order of appearance in algorithms 1 and 2:

- initPi: Initialise the agent's policy to some high-entropy state to encourage exploration.
- initTree: Initialise the tree with a single leaf ($m = 1$).
- rlOneEp: Run one episode of reinforcement learning on the current reward function, using any conventional online RL algorithm suitable for continuous action spaces.

- `append`: Append an element to a list or a row to a matrix.
- `computeN`: Use $\phi$ to compute the **n** vector for a trajectory $\tau$.
- `computeW`: Compute the un-normalised weighting matrix using equation 13 of the main paper, as well as the recency condition for the online setting.
- `sample`: Sample pairs from $\mathcal{T}$ with probabilities $\Psi$.
- `getPreferenceLabel`: Query the human (or synthetic oracle) to obtain a preference label for trajectory pair $\tau_i, \tau_j$.
- `ARow`: Given trajectory pair indices $i, j$, construct a new row for the $A$ matrix as described in section 4.1 of the main paper.
- `numLeaves`: Count the leaves of the tree.
- `RSS`: Perform the residual sum of squares calculation given in equation 10 of the main paper.
- `computeNxdc`: Compute the $2 \times n$ matrix $N^{[xdc]}$ by counting the number of timesteps each trajectory in $\mathcal{T}$ spends in two children of the $x$th leaf of $\phi$, where the children are created by splitting at threshold $c$ along dimension $d$.
- `argmax`: Return the index of the maximum of a list.
- `applySplit`: Update the tree by splitting the $x$th leaf at threshold $c$ along dimension $d$.
- `splitReplace`: Replace the $x$th row of a matrix with a provided 2-row matrix, thereby increasing the number of rows by 1.
- `getLastSplit`: Return the index $x$ of the most-recently split leaf in a tree.
- `applyMerge`: Update the tree by merging the $x$th and $x + 1$th leaves into a single one.
- `mergeReplace`: Replace the $x$ and $x + 1$th elements of a list (respectively, rows of a matrix) with a single provided element (row), thereby decreasing the number of elements (rows) by 1.
- `vecToDiag`: Convert an $m$-dimensional vector to an $m \times m$ diagonal matrix.
- `argmin`: Return the index of the minimum of a list.

*Offline setting as a special case.* Algorithm 1 subsumes both online and offline PbRL settings, with the latter recovered by setting $f_l = n_{\max} = |\mathcal{T}_{\text{off}}| \geq 2$ and $n_{\text{post fix}} > 0$. The effect of this is to ensure lines 8-11 are skipped (preventing the collection of online data) and lines 14-33 are run (creating one large label batch of size $k_b = k_{\max}$). The condition on line 35 is then triggered, breaking the outer loop on the first iteration, and RL training is run for $n_{\text{post fix}}$ episodes using a fixed reward function (lines 40-42).

**Algorithm 1:** Interpretable preference-based RL with a tree-structured reward function.

1 **Input:** Offline trajectory dataset $\mathcal{T}_{\text{off}}$, hyperparameters $n_{\max}, f_l, \lambda, k_{\max}, f_u, m_{\max}, D_s, D_a, \alpha, n_{\text{post fix}}$
2 **Output:** Final agent policy $\pi$
3
4 /* Initialise policy, tree and persistent data structures */
5 $\pi \leftarrow \text{initPi}(); \phi \leftarrow \text{initTree}(); \mathcal{T} = \mathcal{T}_{\text{off}}; \mathbf{r} \leftarrow [0]; \Sigma \leftarrow [0]; \mathcal{P} \leftarrow \{\}; A \leftarrow []; \mathbf{y} \leftarrow []; N \leftarrow []; b \leftarrow 0; \text{done} \leftarrow \textit{False};$
6 **while** *done = False* **do**
7     **if** $|\mathcal{T}| < n_{max}$ **then**
8         /* Run one episode of RL using latest reward function and store trajectory */
9         $\tau, \pi \leftarrow \text{rlOneEp}(\pi, \phi, \mathbf{r});$
10         $\mathcal{T} \leftarrow \text{append}(\mathcal{T}, \tau);$
11         $N \leftarrow \text{append}(N^\top, \text{computeN}(\tau, \phi))^\top;$     /* Double-transpose operation appends a column to N */
12     **end**
13     **if** $|\mathcal{T}| \% f_l = 0$ **then**
14         /* Obtain a batch of preference labels */
15         $b \leftarrow b + 1;$
16         $\mathbf{u} \leftarrow N^\top \mathbf{r} + \lambda \, \text{diag}(N^\top \Sigma N)^{\frac{1}{2}};$     /* Optimistic fitness estimates (paper eqn 12) */
17         $k_b = \text{round}\left(k_{\max} \frac{f_l(2b-1)-1}{\frac{n_{\max}}{f_l}(n_{\max}-1)}\right);$     /* Batch size (paper eqn 14) */
18         **for** $k \in \{1..k_b\}$ **do**
19             $W \leftarrow \text{computeW}(\mathcal{P}, \mathbf{u}, f_l, b);$   /* Un-normalised weighting matrix (paper eqn 13 + recency condition) */
20             **if** $\sum_{ij} W_{ij} > 0$ **then**
21                 /* Sample pair and obtain label, unless all pairs have been sampled already (sum of W = 0) */
22                 $\tau_i, \tau_j \leftarrow \text{sample}(\mathcal{T}, W/\sum_{ij} W_{ij});$
23                 $y_{ij} \leftarrow \text{getPreferenceLabel}(\tau_i, \tau_j);$
24                 $\mathcal{P} \leftarrow \mathcal{P} \cup \{\tau_i, \tau_j\};$
25                 $A \leftarrow \text{append}(A, \text{ARow}(i, j));$
26                 $\mathbf{y} \leftarrow \text{append}(\mathbf{y}, y_{ij});$
27             **end**
28             **if** $k \% f_u = 0$ or $k = k_b$ **then**
29                 /* Update reward function periodically, and at end of batch */
30                 $\phi, \mathbf{r}, \Sigma, N \leftarrow \text{updateRewardFunction}(\phi, \mathcal{T}, \mathcal{P}, N, A, \mathbf{y}, m_{\max}, D_s, D_a, \alpha);$     /* (algorithm 2) */
31                 $\mathbf{u} \leftarrow N^\top \mathbf{r} + \lambda \, \text{diag}(N^\top \Sigma N)^{\frac{1}{2}};$   /* Recompute optimistic fitness estimates (paper eqn 12) */
32             **end**
33         **end**
34     **end**
35     **if** $|\mathcal{T}| = n_{max}$ **then**
36         done = *True*;     /* Fix reward function once n_max is reached */
37     **end**
38 **end**
39 /* Run remaining RL episodes using fixed reward function */
40 **for** $i \in \{1..n_{post\,fix}\}$ **do**
41     $\_, \pi \leftarrow \text{rlOneEp}(\pi, \phi, \mathbf{r});$
42 **end**

**Algorithm 2:** updateRewardFunction subfunction.

1　**Input:** $\phi, \mathcal{T}, \mathcal{P}, N, A, \mathbf{y}, m_{\max}, D_s, D_a, \alpha$
2　**Output:** Feature function $\phi$, reward components $\mathbf{r}, \Sigma$, feature matrix $N$
3
4　/* Compute trajectory-level fitness estimates c */
5　$\tilde{\boldsymbol{\mu}} \leftarrow (A^\top A)^{-1} A^\top \Phi^{-1}(\mathbf{y})$ ;　　　　　　　　　　　　　　　　　　　　　　　　　　　　　　　　　　　　　/* (paper eqn 8) */
6　$m \leftarrow \mathsf{numLeaves}(\phi)$;
7　/* Grow tree to maximum size */
8　**while** $m < m_{max}$ **do**
9　　$Q \leftarrow []; \mathcal{X} \leftarrow []$;
10　　/* Iterate through leaves x, splitting dimensions d and thresholds c */
11　　**for** $x \in \{1..m\}$ **do**
12　　　$\text{rss}_{\text{par}} \leftarrow \mathsf{RSS}(N_x)$ ;　　　　　　　　　　　　/* Residual sum of squares (RSS) for leaf x (paper eqn 10) */
13　　　**for** $d = \{1..D_s + D_a\}$ **do**
14　　　　**for** $c \in \{(s,a)_d, \forall (s,a) \in \tau, \forall \tau \in \mathcal{P}\}$ **do**
15　　　　　$N^{[xdc]} \leftarrow \mathsf{computeNxdc}(\mathcal{T}, \phi, x, d, c)$;
16　　　　　$Q \leftarrow \mathsf{append}(Q, \text{rss}_{\text{par}} - \mathsf{RSS}(N_1^{[xdc]}) - \mathsf{RSS}(N_2^{[xdc]})$ ; /* Quality is reduction in RSS (paper eqn 11) */
17　　　　　$\mathcal{X} \leftarrow \mathsf{append}(\mathcal{X}, (x, d, c, N^{[xdc]}))$ ;　　　　　　　　　　　/* Store split details for retrieval later */
18　　　　**end**
19　　　**end**
20　　**end**
21　　/* Find and apply best split, growing tree */
22　　$x, d, c, N^{[xdc]} \leftarrow \mathcal{X}[\mathsf{argmax}(Q)]$;
23　　$\phi \leftarrow \mathsf{applySplit}(\phi, x, d, c)$;
24　　$m \leftarrow m + 1$;
25　　$N \leftarrow \mathsf{splitReplace}(N, x, N^{[xdc]})$;
26　**end**
27　$\mathbf{r} \leftarrow []; \mathbf{v} \leftarrow []$;
28　/* Compute reward component means and variances */
29　**for** $x \in \{1..m\}$ **do**
30　　$\mathbf{r} \leftarrow \mathsf{append}(\mathbf{r}, \frac{\sum_{\tau_i \in \mathcal{P}} \frac{N_{xi}}{T} \tilde{\mu}_i}{\sum_{\tau_i \in \mathcal{P}} N_{xi}})$ ;　　　　　　　　　　　　　　　　　　　　　　　　/* Mean (paper eqn 9) */
31　　$\mathbf{v} \leftarrow \mathsf{append}(\mathbf{v}, \frac{\mathsf{RSS}(N_x)}{\sum_{\tau_i \in \mathcal{P}} N_{xi}})$ ;　　　　　　　　　　　　　　　　　　　　　　　　　　/* Variance (paper eqn 10) */
32　**end**
33　$\mathcal{L} \leftarrow []; \mathcal{Z} \leftarrow []$;
34　/* Prune tree back to root by iteratively undoing the most recent split */
35　**while** $m > 1$ **do**
36　　$x \leftarrow \mathsf{getLastSplit}(\phi)$;
37　　$\phi \leftarrow \mathsf{applyMerge}(\phi, x)$;
38　　$m \leftarrow m - 1$;
39　　/* Replace xth and x+1th rows/entries of N, r and v with values for merged leaf */
40　　$N \leftarrow \mathsf{mergeReplace}(N, x, N_x + N_{x+1})$ ;　　/* Conservation: timesteps in merged leaf is sum of two children */
41　　$\mathbf{r} \leftarrow \mathsf{mergeReplace}(\mathbf{r}, x, \frac{\sum_{\tau_i \in \mathcal{P}} \frac{N_{xi}}{T} \tilde{\mu}_i}{\sum_{\tau_i \in \mathcal{P}} N_{xi}})$;
42　　$\mathbf{v} \leftarrow \mathsf{mergeReplace}(\mathbf{v}, x, \frac{\mathsf{RSS}(N_x)}{\sum_{\tau_i \in \mathcal{P}} N_{xi}})$;
43　　$\Sigma \leftarrow \mathsf{vecToDiag}(\mathbf{v})$;
44　　/* Tree quality is labelling loss (paper eqn 7) + complexity regularisation */
45　　$\mathcal{L} \leftarrow \mathsf{append}(\mathcal{L}, \left[\Phi^{-1}(\mathbf{y}) - (\mathsf{diag}(NA^\top \Sigma AN^\top)^{-\frac{1}{2}})^\top AN^\top \mathbf{r}\right]^2 + \alpha m)$;
46　　$\mathcal{Z} \leftarrow \mathsf{append}(\mathcal{Z}, (\phi, \mathbf{r}, \Sigma, N))$ ;　　/* Store current state of tree and reward components for retrieval later */
47　**end**
48　/* Find and return best tree */
49　$\phi, \mathbf{r}, \Sigma, N \leftarrow \mathcal{Z}[\mathsf{argmin}(\mathcal{L})]$;

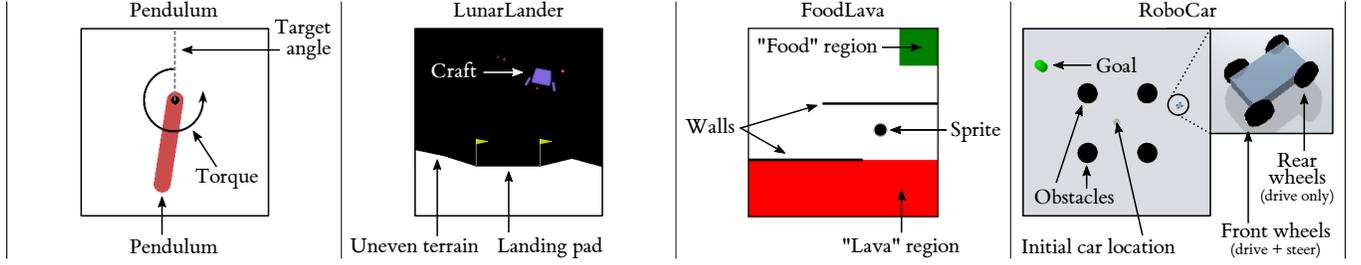

Figure 1: The four RL environments used in experiments.

## APPENDIX E: EXPERIMENTAL DETAILS

### Environments

Our experiments were conducted in four episodic RL environments with Euclidean state-action spaces, all implemented in Python using OpenAI Gym [4]. Figure 1 contains an annotated visualisation of each environment.

*Pendulum-v0.* A built-in component of the Gym library and a classic feedback control problem. The task is to swing an inverted pendulum upright and hold it there for as long as possible up to a time limit of $T = 200$. The state dimensions are the cosine and sine of the pendulum's angle from upright $\theta$ and its angular velocity $\dot{\theta}$ ($D_s = 3$), and the pendulum is initialised in a random position. The action is a torque $u$ on the rotational joint ($D_a = 1$). The default reward function is $R(s_t, a_t) = \theta_t^2 + 0.1\dot{\theta}_t^2 + 0.001u_t^2$.

*LunarLanderContinuous-v2.* Another built-in Gym environment, in which the objective is to guide an aerial craft to a gentle landing on a landing pad surrounded by uneven terrain. We set a time limit of $T = 300$. The state dimensions are the craft's horiontal and vertical positions $x, y$ and velocities $v^x, v^y$, its angle from vertical $\theta$ and angular velocity $\dot{\theta}$, and two binary contact detectors $c^l, c^r$ indicating whether the left and right landing legs are in contact with the ground ($D_s = 8$). The craft is initialised in a narrow zone above the landing pad, with slightly-randomised angle and velocities. The action is a pair of throttle values for two engines: main $u^m$ and side $u^s$ ($D_a = 2$). For most timesteps, the default reward function gives $R(s_t, a_t) = \text{shaping}_t - \text{shaping}_{t-1}$,[2] where $\text{shaping}_t =$

$$-100\left(\sqrt{x_t^2 + y_t^2} + \sqrt{(v_t^x)^2 + (v_t^y)^2} + |\theta|\right) + 10\left(c_t^l + c_t^r\right).$$

In addition, a one-off reward of $+100$ is given if the craft successfully lands on the pad, and $-100$ is given if it crashes or drifts out-of-bounds ($|x| \geq 1$). The distinction between a landing and a crash is based on a force analysis in an underlying rigid-body physics simulation, whose details are undocumented in the provided open source code. To convert LunarLander into a fixed-length episodic task, we disabled a default condition that terminates the episode immediately after a landing, crash or out-of-bounds event.

*FoodLava.* A simple holonomic navigation task created by ourselves to enable easy visualisation of trajectories for debugging during code development. The objective is to control a circular black

[2]Note that this expression abuses notation because it is dependent on the shaping quantity from the previous timestep. The fact that reward depends on more than the current state-action pair means that strictly speaking, LunarLander is a partially-observable MDP.

sprite to escape or avoid a red "lava" region and move to the green "food" while navigating around a pair of walls (black lines). The time limit is $T = 200$. The state dimensions are the sprite's horizontal and vertical positions $x, y$ ($D_s = 2$) and are initialised randomly within the boundaries of the maze layout ($[0, 10]^2$). The action dimensions are horizontal and vertical velocities $v^x, v^y$ ($D_a = 2$), which are clipped if the resultant motion vector would intersect a wall or external boundary. The default reward function is

$$R(s_t, a_t) = \begin{cases} 1 & \text{if } x_t \geq 8 \land y_t \geq 8 & \text{("food" region)} \\ -1 & \text{if } x_t \leq 3 & \text{("lava" region)} \\ 0 & \text{otherwise.} \end{cases}$$

*RoboCar.* Also created by ourselves using the PyBullet 3D physics simulator [6], the task in this environment is to drive a four-wheeled car to a green goal object while avoiding four black obstacles. The time limit is $T = 200$. The state dimensions are the coordinates of the car's centroid $x, y$, the cosine and sine of its orientation $\theta$, its velocity components $v^x, v^y$, the distance and bearing in radians to the goal $d, \beta$ ($\beta = 0$ when facing the goal) and a binary indicator of contact with an obstacle $c^o$ ($D_s = 9$). The car is initialised at $(x, y, \theta, v^x, v^y) = (0, 0, 0, 0, 0)$ and the obstacles are always the same, but the goal location is randomised on each episode. The action is a throttle $u^t$ and steering angle $u^s$, which are applied subject to limits and a simple model of drag and mechanical resistance ($D_a = 2$). The default reward function is $R(s_t, a_t) = -0.05d_t - 0.1c_t^o + [d_t < 2]$, where $[\cdot]$ is Iverson bracket notation; this final term adds a reward of $+1$ if the car is within a radius of 2 from the goal.

### Common Setup and Parameters

*Feedback budget.* Since our survey (offline human experiment) received approximately 60 respondents, each of which provided 10 preference labels per environment, we used a feedback budget of $k_{\max} = 600$ for all other experiments. This figure enabled direct comparison of our algorithm's performance across environments and experiment types, and also reflects what we see as a reasonable demand on human labour (on the order of 1 hour, assuming $\approx 10$ labels per minute).

*Preference label noise.* For the noise parameter used to prevent extreme probabilities, we used $\varepsilon = 0.1$, which effectively set a minimum 10% chance that the higher-fitness trajectory was erroneously preferred. A 10% random error rate matches that used in prior PbRL work by Christiano et al. [5].

*Optimism parameter for UCB sampling.* Throughout all experiments we used $\lambda = 2$, meaning the optimistic fitness estimates $\mathbf{u}$ were 2 standard deviations above the means $\boldsymbol{\mu}$.

*Tree growth parameters.* We used a maximum tree size of $m_{\max} = 100$ and a complexity regularisation parameter of $\alpha = 0.01$. We found that the final reward structure was sometimes quite sensitive to the latter, which could easily have been tuned for each environment and experimental context. For the sake of simplicity and to avoid cherry-picking, we identified this single value as one that provided good performance across all four environments.

*RL algorithm.* For all RL agents – both pilot and PbRL – we used the soft-actor critic algorithm [7] with discount factor $\gamma = 0.99$, learning rates of $1e^{-4}$ and $1e^{-3}$ for the policy and value networks respectively, an entropy regularisation coefficient of 0.2, and an interpolation factor of 0.99 for Polyak averaging of the target networks. All networks had two hidden layers of 256 units each. The replay buffer capacity $B$, minibatch size $M$ and total number of training episodes $E$ were independently selected for each environment after an informal search, and held constant across all experiments:

|   | Pendulum | LunarLander | FoodLava | RoboCar |
|---|----------|-------------|----------|---------|
| $B$ | 5000 | 20000 | 20000 | 40000 |
| $M$ | 32 | 64 | 128 | 64 |
| $E$ | 200 | 200 | 400 | 1000 |

Note that for the offline experiments, $E = n_{\max} = n_{\text{post fix}}$, because these involved first training a pilot agent to create an offline trajectory dataset ($E = |\mathcal{T}_{\text{off}}| = n_{\max}$), then training a PbRL agent on the resultant fixed reward function ($E = n_{\text{post fix}}$). For the online experiments, $E = n_{\max} + n_{\text{post fix}}$, because the training of the PbRL agent consisted of a phase of online preference gathering and reward modification (up to $n_{\max}$) followed by a phase of continued training on a fixed reward function (another $n_{\text{post fix}}$ episodes).

*Repeated experimental runs.* In all experiments other than the most labour-intensive online study with human feedback, we trained 5 PbRL agents for each environment, with mean, minimum and maximum performance shown on the learning curve plots (main paper figures 2-4). In the offline setting, all repeats used the same learnt reward function, so the variation reflected the stochasticity of the RL process only. In the online setting, a reward function was constructed from scratch during each agent's training. Since this process naturally differed slightly between runs, it provided an additional source of variation compared with the offline experiments.

### Offline with Oracle Feedback

*Oracle implementation.* To implement the oracles for automated preference labelling, we directly queried the environment's default reward function for each state-action pair in the two trajectories $\tau_i, \tau_j$, summed over the trajectory lengths, and returned $y_{ij} = 1 - \varepsilon = 0.9$ if the sum for $\tau_i$ exceeded that for $\tau_j$, $y_{ij} = \varepsilon = 0.1$ if vice versa, and $y_{ij} = 0.5$ if the sums were equal. While adding more than these three preference levels would increase the information content of each label, it would have required further assumptions about the mapping from fitness differences to preferences, and we found that the algorithm performed well without this added complexity.

*Update frequency.* As described in Appendix D, preference elicitation in the offline setting consists of a single batch of $k_{\max}$ samples, but the batch can be paused every $f_u$ samples to refine the tree structure, reward components and sampling distribution. In our offline oracle experiment we used $f_u = 60$. Given that $k_{\max} = 600$, this meant that a total of 10 updates were completed during the feedback process.

### Offline with Human Feedback

*Survey administration.* Our survey was run via Google Forms, with trajectory pairs displayed in a separately-hosted web application; this was required to enable each participant to receive a randomised set of pairs. As outlined in Appendix A, pairs were displayed side-by-side as infinitely-looping videos ($\tau_i$ on the left, $\tau_j$ on the right). Videos were generated by enabling OpenAI Gym's monitor wrapper during the training of the pilot agents, and appeared visually as animated versions of the images in figure 1 above (for RoboCar, the bird's eye view was used). Survey participants gave ratings on a discretised 0-10 scale, with 0 corresponding to the strongest preference for $\tau_i$ (interpreted as $y_{ij} = 1 - \varepsilon = 0.9$), vice versa for 10, and a linear interpolation for intermediate values (so a rating of 5 corresponded to $y_{ij} = 0.5$). Participants were given 10 trajectory pairs to rate for each environment. We publicised our survey among research groups in the fields of computer science and engineering, as well as non-expert personal and professional contacts. Over 1 week of data collection we had 62 respondents of a wide variety of experience levels (see main paper figure 3b), yielding a total preference dataset size of $k_{\max} = 62 \times 10 = 620$ for each environment. Prior to any human experiments being run, our survey design received full university ethics approval.

*Lack of periodic updates.* Since the survey was run via the internet with asynchronous responses from many participants, it would have added significant technical complexity to perform tree and reward component updates on the back-end throughout the 1 week survey period. For this reason, we waited until all responses were gathered before doing a single update step, thereby effectively setting $f_u \geq 620$. Since our algorithm initiates with equal fitness estimates for all trajectories, the result was that pairs were sampled uniform-randomly throughout the survey, subject to the non-identity, duplication and connectivity conditions given in section 4.5 of the main paper. Unfortunately, this somewhat reduced the comparability of these results to those from the offline oracle experiment, where $f_u = 60$. We would likely have attained better performance in this experiment if periodic updates were possible, as it would have enabled the UCB weighting method to work as designed. Developing a more sophisticated pipeline for large-scale user experiments is a clear priority for future work.

### Online with Oracle Feedback

*Oracle implementation.* The oracles were identical to those used in the offline experiment.

*Fixing of reward function.* PbRL agents were trained for a total of $E = n_{\max} + n_{\text{post fix}}$ episodes, where $E$ varied between environments as given in the table above. The training period was divided into two phases of pre- and post-fixing of the reward function as follows:

|  | Pendulum | LunarLander | FoodLava | RoboCar |
|---|---|---|---|---|
| $n_{max}$ | 100 | 100 | 100 | 200 |
| $n_{post\ fix}$ | 100 | 100 | 300 | 800 |

*Batch and update frequencies.* In all environments, we obtained a batch of preference labels every $f_l = 10$ episodes until $n_{max}$ was reached, with batch sizes given by the scheduling equation (main paper equation 14). We found no measurable benefit to performing tree and reward component updates more than once per batch, so set $f_u$ to an arbitrarily large value to disable this feature.

## Online with Human Feedback

*User interface.* Using the OpenCV Python library [3], we constructed a graphical user interface to obtain human preference labels over trajectories generated by a PbRL agent running locally on the same machine. Mirroring the survey used in the offline human experiment, the interface presented trajectory pairs side-by-side as infinitely-looping videos, and user input was constrained to a 0-10 scale (mapped to numerical keys on the keyboard) with 0 and 10 representing maximal preference for $\tau_i$ and $\tau_j$ respectively.

*Participant.* This experiment was run with a single human participant: one of the authors of this paper.

*Fixing of reward function.* As discussed in the main paper, the first run for both FoodLava and LunarLander used the same values of $n_{max}$ and $n_{post\ fix}$ as the oracle-based experiment. After observing the premature convergence phenomenon, we updated to the following values for the second run:

|  | FoodLava | LunarLander |
|---|---|---|
| $n_{max}$ | 300 | 190 |
| $n_{post\ fix}$ | 100 | 10 |

*Batch and update frequencies.* These were unchanged from the online experiment with oracle feedback.

## APPENDIX F: CAUSAL CONFUSION OF BEARING INFORMATION IN ROBOCAR

In the failure case encountered when learning a reward function for RoboCar from offline human feedback (the survey responses), we find that part of the misalignment was due to a counterintuitive treatment of the bearing information $\beta$ when the car was in the $y \in [-1.64, 1.68]$ corridor with a distance-to-goal $d \geq 5.84$. The offending subtree is shown in figure 2 below.

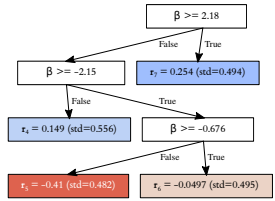

Figure 2: Subtree containing $\beta$-based splits.

The first two splits here created positive reward components for $\beta \geq 2.18$ or $\beta < -2.15$, and negative reward otherwise. The agent was thus rewarded for facing *away* from its goal, which is heavily misaligned. The near-exact symmetry of these two splits implies that they were due to a reliable feature of the task rather than a statistical fluke, and our diagnosis is as follows.

In RoboCar, the agent must learn to navigate the car around four black obstacles in order to reach the goal. Thus, in the pilot trajectory dataset $\mathcal{T}$, many successful goal-reaching trajectories featured the car initially moving away from the goal to bypass an obstacle, before turning to face the goal and moving rapidly towards it. As a result, a large proportion of timesteps belonging to trajectories with high inferred fitness fell outside of the $\beta \in [-2.15, 2.18]$ region. Meanwhile, many unsuccessful trajectories involved the car driving directly at the goal, and becoming stuck against an obstacle for many timesteps in a row. Thus, if the car was *both* far from the goal (i.e. $d \geq 5.84$) *and* facing towards it ($\beta \in [-2.15, 2.18]$), it was statistically more likely that this timestep was a member of a low-fitness trajectory than a high-fitness one. The exact opposite was true for timesteps spent facing away from the goal. This led our algorithm to fall foul of causal confusion by creating a tree that rewarded facing away from the goal; a circumstance which merely correlated with high fitness rather actively than driving it.

An aligned solution to this problem would be to instead introduce a split on the obstacle contact indicator variable $c^o$, creating a reward component that penalised collisions directly. It is not entirely clear why this option was not taken, but it may be that the human survey participants were not sufficiently consistent in the penalisation of obstacle contact in their responses. What we can be confident about is that an online learning setup would help to reduce the likelihood of such a confusion persisting, as it would provide an opportunity for the human(s) to reactively penalise early examples of behaviour that resulted from it.

We have not been able to meaningfully diagnose of the final, asymmetric split at $\beta = -0.676$, which suggests that this is likely due to a random statistical imbalance in the training dataset $\mathcal{T}$.



\end{document}